\documentclass[11pt]{article}
\pdfoutput=1  % arXiv: force pdfLaTeX

\usepackage[utf8]{inputenc}
\usepackage[T1]{fontenc}
\usepackage{lmodern}
\usepackage[margin=1in]{geometry}
\usepackage{microtype}
\usepackage{amsmath,amssymb}
\usepackage{adjustbox}  % loads graphicx; max width=\linewidth keeps wide figures inside the text block
\usepackage{needspace}  % keep a heading+paragraph together so floats don't split it
\usepackage{xcolor}
\usepackage{booktabs}
\usepackage{array}
\usepackage{enumitem}
\usepackage{listings}
\usepackage{float}
\usepackage{placeins}
\usepackage{caption}
\usepackage{subcaption}

\usepackage{tikz}
\usetikzlibrary{arrows.meta,positioning,calc,shapes.geometric,fit,backgrounds,decorations.pathreplacing}
\usepackage[numbers,sort&compress]{natbib}
\usepackage[colorlinks=true,linkcolor=blue!55!black,citecolor=teal!60!black,urlcolor=violet!70!black]{hyperref}
\usepackage{authblk}

\definecolor{lapblue}{HTML}{1F5FA6}
\definecolor{lapteal}{HTML}{0E7C7B}
\definecolor{lapamber}{HTML}{C77A14}
\definecolor{lapred}{HTML}{B3261E}
\definecolor{lapgray}{HTML}{4A4A4A}
\definecolor{laplight}{HTML}{EAF1F8}
\definecolor{laplight2}{HTML}{E5F2F1}
\definecolor{codebg}{HTML}{F6F8FA}

\lstdefinelanguage{json}{
  basicstyle=\ttfamily\scriptsize,
  showstringspaces=false,
  breaklines=true,
  breakatwhitespace=true,
  frame=single,
  rulecolor=\color{lapgray!30},
  backgroundcolor=\color{codebg},
  literate=
   *{:}{{{\color{lapgray}{:}}}}{1}
    {,}{{{\color{lapgray}{,}}}}{1}
    {\{}{{{\color{lapblue}{\{}}}}{1}
    {\}}{{{\color{lapblue}{\}}}}}{1}
    {[}{{{\color{lapblue}{[}}}}{1}
    {]}{{{\color{lapblue}{]}}}}{1},
}
\tikzset{
  box/.style={rounded corners=2pt,draw=lapgray!60,thick,fill=white,align=center,
    inner sep=4pt,font=\small},
  layer/.style={rounded corners=2pt,draw=lapblue!50,thick,fill=laplight,align=center,
    minimum height=8mm,font=\small,inner sep=4pt},
  role/.style={rounded corners=3pt,draw=lapteal!70,very thick,fill=laplight2,align=center,
    font=\small\bfseries,inner sep=5pt},
  state/.style={ellipse,draw=lapblue!70,thick,fill=laplight,align=center,font=\footnotesize,
    inner sep=2pt,minimum width=18mm},
  sstate/.style={ellipse,draw=lapamber!80,thick,fill=lapamber!12,align=center,font=\footnotesize,
    inner sep=2pt,minimum width=18mm},
  term/.style={ellipse,draw=lapgray!70,thick,fill=lapgray!10,align=center,font=\footnotesize,
    inner sep=2pt,minimum width=16mm},
  flow/.style={-{Stealth[length=2.2mm]},thick,lapgray!80,font=\scriptsize},
  lifeline/.style={lapgray!40,thick},
  msg/.style={-{Stealth[length=2mm]},thick,lapblue!80,font=\scriptsize},
  rmsg/.style={-{Stealth[length=2mm]},thick,lapteal!80,dashed,font=\scriptsize},
}

\title{\textbf{LAP: An Agent-to-Instrument Protocol\\ for Autonomous Science}}
\author[1]{Linwu Zhu}
\author[1]{Liqiang Gao}
\author[1]{Yan Chen}
\author[1]{Dan Zhu}
\author[1]{Jian Huang}
\affil[1]{Shiyanjia Lab \\ \texttt{\{zhulinwu, gaoliqiang, chenyan, zhudan, huangjian\}@shiyanjia.com}}
\date{\today}

\begin{document}
\maketitle

\begin{abstract}
\noindent
Autonomous science is moving from demonstration to infrastructure. Large language
model agents now plan experiments, and self-driving laboratories execute them.
Yet every such system rebuilds the link between the reasoning agent and the physical
instrument from scratch, against fragmented vendor SDKs and standards built for
deterministic software clients rather than probabilistic, goal-directed agents. Recent
agent-interoperability protocols clarify two of the three edges of an agentic
ecosystem (Anthropic's Model Context Protocol (MCP) standardizes the
\emph{agent-to-tool} edge, and Google's Agent2Agent (A2A) the \emph{agent-to-agent}
edge), but neither models the \emph{agent-to-instrument} edge, where operations are
stateful, safety-critical, exclusively owned, physically embodied, and produce
measurements with units, calibration, and uncertainty. We present the \textbf{Lab Agent
Protocol (LAP)}, a protocol design that fills this gap. LAP retains A2A's peer-to-peer,
discovery-first, task-lifecycle structure and adds four physical-world primitives:
(i) the \emph{InstrumentCard}, a signed capability and physical-limit description;
(ii) first-class \emph{reservation} for exclusive instrument and sample locking;
(iii) a \emph{safety-fence handshake} with operator-confirmation tokens cryptographically
bound to a specific task and its parameters, gating hazardous and irreversible
operations; and (iv) a \emph{MeasurementResult} schema that makes every result physically
typed (QUDT/UCUM), calibration-anchored, uncertainty-bearing, and reproducible by
construction. We specify roles, a six-layer architecture, the JSON-RPC method set, the
task and safety state machines, the error model, and cross-laboratory federation, and
walk a closed-loop autonomous campaign through the protocol end-to-end. LAP is
transport-compatible with the A2A/MCP ecosystem and encapsulates rather than replaces
existing device standards such as SiLA\,2 and OPC-UA.
\end{abstract}

\vspace{0.5em}
\noindent\textbf{Keywords:} autonomous science, self-driving laboratories, agent
protocols, AI agents, laboratory automation, instrument interoperability, A2A, MCP,
agentic AI.

\vspace{0.6em}
\noindent\textbf{Status of this document:} This is a design specification (LAP~v0.1)
intended to seed an open, community-built standard; it has no normative or implemented
status yet, and its benefits are argued architecturally rather than empirically validated.

\clearpage
\tableofcontents
\clearpage

% ============================================================
\section{Introduction}
\label{sec:intro}
\subsection{The bottleneck of autonomous science}
A decade of progress in laboratory robotics and, more recently, in large language model
(LLM) agents has made a once-speculative idea concrete: laboratories that design,
execute, and interpret their own experiments. Berkeley's A-Lab synthesized novel
inorganic materials around the clock with no human in the loop~\citep{alab2023};
GPT-4-driven systems such as Coscientist planned and ran chemical reactions on a cloud
laboratory~\citep{coscientist2023}; tool-augmented chemistry agents such as
ChemCrow chained reasoning over many instruments~\citep{chemcrow2024}; and a mobile
robotic chemist ran a multi-day optimization campaign autonomously~\citep{liverpool2020}.
The reasoning layer of autonomous science (hypothesis generation, experiment design,
active learning, result interpretation) is improving rapidly and is increasingly
delivered by general-purpose agent frameworks.

The layer that is \emph{not} improving at the same rate is the link between the
reasoning agent and the physical instrument. In every system cited above, that link is
bespoke. A-Lab's orchestrator drives twenty-eight devices across sixteen device types,
each through a custom-written adapter~\citep{alabos2024}. Coscientist must parse vendor
documentation at run time to control each new instrument~\citep{coscientist2023}.
ChemCrow integrates its tools through individually engineered wrappers~\citep{chemcrow2024}.
Recent reviews are consistent: self-driving laboratories operate in silos, instruments
expose diverse vendor APIs with inconsistent command interfaces, telemetry formats, and
safety semantics, and integration is handled through fragile hacks rather than robust
interfaces~\citep{sdl2026,eac2025,nist2024}. Each new laboratory, and often each new
instrument, re-implements the same plumbing. This is precisely the kind of $O(n\times m)$
integration problem ($n$ agents against $m$ instruments) that a protocol is meant to
collapse into $O(n+m)$.

\subsection{Three edges, two protocols}
The broader AI community has, in the span of roughly a year, converged on protocols for
exactly this class of problem, but for software rather than for matter. Anthropic's
\emph{Model Context Protocol} (MCP) standardizes how a single agent reaches tools,
data, and services: the \emph{agent-to-tool} edge~\citep{mcp2024}. Google's
\emph{Agent2Agent} (A2A) protocol standardizes how autonomous agents discover, delegate
to, and coordinate with one another as peers: the \emph{agent-to-agent}
edge~\citep{a2a2025}. Both have been donated to neutral governance under the Linux
Foundation and are accumulating large ecosystems. They are complementary: MCP is the
vertical edge from an agent down to its tools; A2A is the horizontal edge between agents.

A scientific agent ecosystem has a third edge that neither protocol addresses: the
\emph{agent-to-instrument} edge. An instrument is not a stateless tool call and not a
peer software agent. It is a stateful, physically embodied resource. Its operations
consume real reagents and irreplaceable samples; they take minutes to hours; they can
injure people or destroy equipment; they require exclusive access while running; and
they produce results that are only meaningful when accompanied by physical units,
a calibration reference, and an uncertainty. None of these properties is expressible in
an MCP tool schema or an A2A AgentCard. Existing laboratory standards
(SiLA\,2, OPC-UA, SCPI, AnIML, Allotrope) address parts of the device and data layers,
but all of them assume a deterministic software client that knows in advance exactly
which command to issue, not a probabilistic agent that reasons from a natural-language
goal and must \emph{discover} what an instrument can do, \emph{negotiate} whether an
action is safe, and \emph{interpret} a physically typed result~\citep{sila2,opcua}.

\subsection{Contributions}
This technical report presents the \textbf{Lab Agent Protocol (LAP)}, a complete design
for the missing agent-to-instrument edge. Our specific contributions are:
\begin{itemize}[leftmargin=1.4em,itemsep=2pt]
\item \textbf{A problem framing} (\autoref{sec:intro}, \autoref{sec:overview}) that
  locates instrument control as the third edge of the agentic ecosystem and derives the
  four physical-world primitives any such protocol must provide.
\item \textbf{The InstrumentCard} (\autoref{sec:protocol}), a signed, JSON-LD capability
  and physical-limit description that extends the A2A AgentCard and the W3C Web of Things
  Thing Description with instrument class, per-capability safety class, reversibility,
  physical limits, calibration state, and natural-language intent grounding.
\item \textbf{A complete interaction layer}: a JSON-RPC\,2.0 method set, a task lifecycle
  state machine that augments A2A's eight states with physical states
  (\emph{safety-hold}, \emph{paused-fault}, \emph{sample-wait}), first-class
  \emph{reservation} leases for exclusive locking, and a structured error model
  (\autoref{sec:protocol}).
\item \textbf{A safety-fence handshake} (\autoref{sec:protocol}) built on a four-level
  safety classification (S0--S3) and operator-confirmation tokens cryptographically
  bound to a specific task and its parameter hash, so that an autonomous agent's
  authority over hazardous or irreversible actions is explicit, bounded, and auditable.
\item \textbf{The MeasurementResult schema} (\autoref{sec:protocol}), which makes every
  result physically typed via QUDT/UCUM, calibration-anchored, uncertainty-bearing, and
  accompanied by a reproducibility manifest and a signature.
\item \textbf{A federation and discovery design} (\autoref{sec:federation}) for
  cross-laboratory capability search, scoped credential grants, and physical-sample
  chain-of-custody, together with an end-to-end worked example of a closed-loop
  autonomous campaign.
\end{itemize}

\noindent
LAP is not a new transport or data format. It reuses JSON-RPC, JSON Schema,
QUDT/UCUM, W3C WoT Thing Descriptions, and A2A's discovery and task patterns, and it
encapsulates rather than competes with device standards such as SiLA\,2, which live
\emph{below} LAP inside an Instrument Agent. Its novelty is the combination of the
four physical-world primitives into one protocol for the agent-to-instrument edge, which,
to our knowledge, no existing protocol provides. This report is a
design and specification, not yet a large-scale implementation or empirical evaluation;
we discuss this limitation candidly in \autoref{sec:discussion}.

% ============================================================
\section{Background and Related Work}
\label{sec:related}

\subsection{Autonomous and Self-Driving Laboratories}
\label{sec:related:sdl}

The past five years have produced a series of demonstrations that AI-driven autonomous
experimentation is physically feasible.
\citet{alab2023} introduced the \textbf{A-Lab} at Lawrence Berkeley National Laboratory,
a closed-loop system that combined density-functional phase-stability data from the
Materials Project, natural-language synthesis-recipe extraction, and active-learning
optimization to synthesize 36 out of 57 target inorganic compounds in 17 days of
unattended operation.
Critically, its orchestration framework, AlabOS, maintains 28 device instances spanning
16 device types, each requiring a hand-coded protocol adapter (MODBUS, TCP/IP, serial,
HTTP, XML-RPC)~\citep{alabos2024}, illustrating concretely the bespoke-driver tax that
every self-driving laboratory (SDL) currently pays.

\citet{coscientist2023} demonstrated \textbf{Coscientist}, a multi-LLM agent system that
autonomously designed, planned, and physically executed palladium-catalyzed cross-coupling
reactions by composing internet search, instrument-documentation retrieval, code
generation, and remote calls to the Emerald Cloud Lab's robotic API.
The generalization problem is concrete: for every new instrument the LLM must
parse vendor-specific API documentation at runtime, which scales poorly and
depends on documentation quality rather than a machine-readable capability contract.

\citet{chemcrow2024} augmented GPT-4 with 18 expert-designed chemistry tools spanning
retrosynthesis, safety checks, and literature search.
Each of the 18 tools required a bespoke integration; scaling to hundreds of instruments
demands a standard.

The \textbf{Liverpool mobile robotic chemist}~\citep{liverpool2020} sent a 1.75-metre
robot to navigate a shared chemistry lab autonomously, performing approximately 700
photocatalyst experiments over eight days and identifying a catalyst with 6$\times$
improved efficiency.
A 2024 follow-up~\citep{liverpool2024} demonstrated two cooperating AI-driven robots
making autonomous synthesis and analysis decisions at human-competitive timescales.
Cooper's ``automate the researcher, not the instrument'' strategy sidesteps the driver
problem by using robot manipulators on unmodified equipment, but does not eliminate the driver problem.

The \textbf{Emerald Cloud Lab}~\citep{ecl} is a remote-access laboratory-as-a-service
where scientists submit experiments via a programmatic interface and robotic infrastructure
executes them.
It demonstrates that fully programmable cloud labs are commercially viable, while
remaining a closed, proprietary ecosystem whose instruments are not individually
discoverable or taskable by external agents.

The \textbf{Acceleration Consortium} at the University of Toronto has assembled a network
of over 30 partner SDLs worldwide and produced one of the most extensive demonstrations of
networked autonomous science to date:
\citet{delocalized2024} orchestrated a single AI-managed discovery campaign across
laboratories in Toronto, Vancouver, Glasgow, Illinois, and Fukuoka, synthesizing and
characterizing more than 1,000 organic solid-state laser emitter candidates and
identifying 21 top performers, the first globally networked, delocalized, asynchronous
closed-loop SDL.
The campaign ran on \textbf{ChemOS 2.0}~\citep{chemos2024}, which uses SiLA2 for device
control; yet even this system required bespoke multi-site integration
that does not generalize to a new instrument or a new collaborating lab.

\textbf{Bluesky and Ophyd}~\citep{bluesky,bluesky_paper} (NSLS-II, Brookhaven National Laboratory) constitute the \emph{de facto} experiment-orchestration and hardware-abstraction stack at US synchrotron beamlines, deployed on 26 of 28 NSLS-II beamlines and adopted at the Advanced Photon Source (APS), the Stanford Synchrotron Radiation Lightsource (SSRL), the Canadian Light Source, and Diamond Light Source.
Ophyd provides a uniform Python device-object model over EPICS process variables; Bluesky's RunEngine executes scan plans against those objects and emits structured event streams to a document model.
LAP sits conceptually above this stack: an Instrument Agent wrapping a Bluesky-controlled beamline would expose its capabilities as an InstrumentCard and accept LAP task submissions, while continuing to drive hardware via Ophyd internally.

\textbf{IvoryOS}~\citep{ivoryos2025} (Acceleration Consortium, University of Toronto, \emph{Nature Communications} 2025) is a web-interface SDL orchestrator that automatically generates drag-and-drop graphical workflows from Python hardware-driver APIs, enabling within-institution coordination of SDL components without hand-written integration code.
IvoryOS operates at the UI and orchestration level within a single institution; it does not define a wire protocol, provides no safety-fence or reservation primitive, and was not designed for cross-institutional federation (the scope LAP addresses).

Across all these systems the same structural bottleneck recurs.
\citet{sdl2026} summarize it pointedly: \textit{``SDLs often operate in silos, with
experiment data, metadata, and procedures stored in formats customized for hardware setup
of each laboratory\ldots\ integrating diverse instruments can be cumbersome in the absence
of common interfaces.''}
NIST has launched an explicit program to address what it calls \textit{``fragile hacks
rather than robust interfaces''} when instruments are adapted for AI control, drawing an
analogy to how internet communication standards revolutionized digital technology~\citep{nist2024}.
LAP is a direct response to this identified gap.

\subsection{Laboratory Automation Standards}
\label{sec:related:standards}

Several established standards govern aspects of laboratory instrument communication; none
was designed for the probabilistic, goal-directed AI agent as client.

\textbf{SiLA 2}~\citep{sila2} is the most mature open standard for laboratory instrument
connectivity.
Built on gRPC over HTTP/2 with Protocol Buffers serialization, it defines a
Feature Definition Language (FDL) that provides machine-readable, typed capability
descriptions for instrument commands and observable (streaming) properties, and mandates
mDNS/DNS-SD for plug-and-play discovery.
SiLA 2 excels at deterministic orchestration: a workflow engine can enumerate every
command an instrument accepts, invoke long-running operations with streaming progress, and
discover new instruments on a lab network without configuration.
However, FDL schemas are static and defined at design time; there is no runtime capability
negotiation, no mechanism for expressing physical safety limits or hazard classifications
per capability, no concept of exclusive resource reservation, no structured uncertainty in
results, and no provision for a probabilistic reasoning agent to ground natural-language
intent against what the instrument can actually do in its current state.
The SiLA 2 specification contains no provisions for AI or LLM integration.

\textbf{AnIML}~\citep{animl} (Analytical Information Markup Language, ASTM) is an
XML-based standard for representing analytical chemistry measurement results.
It covers the data-at-rest layer with strong audit-trail and GxP compliance support, and a
2026 OWL 2 ontology extension~\citep{animlontology2025} aligns it semantically with
Allotrope.
AnIML cannot command an instrument or stream live data; it represents completed
measurements, not the control loop.

\textbf{Allotrope/ADF}~\citep{allotrope} combines an HDF5 binary file format with a
richly axiomatized OWL 2 ontology suite (the Allotrope Foundation Ontologies, $\sim$5,000
terms) and SHACL-validated data models for specific analytical techniques.
ADF is the semantically richest standard for capturing what was measured, with what, and
under what conditions; however, it is a data-archival format with no instrument control
semantics, no real-time streaming model, and gated access to full tooling requiring
consortium membership.

\textbf{OPC-UA with LADS}~\citep{opcua,lads} provides a mature, secure industrial
information model that OPC-UA Companion Specification LADS extends with lab-domain
concepts (FunctionalUnit, ProgramManager, ResultType, device state machines).
OPC-UA's heavyweight node-graph architecture and generic information model carry no
lab-specific semantic meaning out of the box; LADS adds domain structure but remains
template-based; there is no mechanism for composing novel procedures at runtime or
expressing the scientific intent behind a measurement.

\textbf{SCPI/VISA}~\citep{scpi} standardizes ASCII command syntax and I/O abstraction for
electronic test and measurement instruments (oscilloscopes, power supplies, signal
generators), but explicitly excludes analytical chemistry and life-science instruments.
SCPI commands are syntactically normalized but semantically opaque; there is no structured
capability discovery, no push-notification model, and no machine-interpretable data types
in responses.

\textbf{EPICS} (Experimental Physics and Industrial Control System)~\citep{epics} is the dominant distributed real-time control framework at particle-accelerator and synchrotron-beamline facilities worldwide, providing a Channel Access network layer over which detector, motion, and vacuum systems publish and subscribe to process variables.
EPICS operates at the device-control layer (analogous to LAP's L0); LAP is designed to sit above it, with an Instrument Agent encapsulating EPICS process variables behind a capability-described, reservation-aware, safety-fenced interface.

\textbf{W3C WoT Thing Description 1.1}~\citep{wottd} is a W3C Recommendation (December 2023) that provides a JSON-LD information model for describing IoT devices: their Properties, Actions, and Events with multiple protocol bindings (HTTP, MQTT, CoAP) and a \texttt{unit} annotation field (recommending QUDT).
LAP's \textbf{InstrumentCard} is explicitly a \emph{profile} of WoT TD~1.1, extending it with laboratory-domain semantics: per-capability safety classification, physical-limit declarations, calibration and uncertainty blocks, reservation profile flags, and the \texttt{intentTags} grounding field.
WoT TD already handles the capability-description layer; LAP's novelty lies instead in the safety-fence handshake with parameter-hash-bound operator tokens, the first-class reservation primitive, and the calibration-and-uncertainty-mandated result schema, none of which WoT TD defines.

\textbf{SLAS/ANSI-SBS microplate geometry standards}~\citep{slas} define the
127.76\,mm $\times$ 85.48\,mm footprint and well-position conventions that allow liquid
handlers, plate readers, and centrifuges from different vendors to physically exchange
plates.
This purely physical layer is a necessary prerequisite for high-throughput automation but
carries no digital, control, or semantic content.

The shared insufficiency of all these standards for AI agents is structural, not incidental:
they assume a \emph{deterministic software orchestration client} that knows in advance
exactly which commands to issue.
WoT TD~1.1 comes closest to the capability-description layer, but no prior standard provides all four primitives as a coherent agent-to-instrument layer: intent grounding (what goal does the agent want to achieve?), safety negotiation with parameter-hash-bound operator tokens (what human authorization is required before irreversible or hazardous actions?), runtime capability discovery anchored to the instrument's current physical state, and calibration-anchored measurement results with propagated uncertainty.

\subsection{Agent Interoperability Protocols}
\label{sec:related:agents}

The 2024--2025 period produced a cluster of open protocols addressing agent-to-agent and
agent-to-tool interoperability.
\citet{ehtesham2025} provide a systematic survey and fifteen-dimension taxonomy of the
four leading protocols.

\textbf{MCP} (Model Context Protocol, Anthropic, November 2024; donated to the Linux
Foundation Agentic AI Foundation in December 2025)~\citep{mcp2024} is a
\emph{vertical} agent-to-tool protocol: a single LLM (the Host) connects through a
stateful Client to a Server that exposes Tools (callable functions with JSON Schema
input/output), Resources (contextual data), and Prompts.
Discovery is through manual host configuration; tools are described as JSON Schema objects
with free-text descriptions.
MCP is hub-and-spoke: the LLM sees all tools but two Servers cannot communicate with each
other, and there is no concept of task delegation, peer-to-peer capability negotiation,
or resource reservation between agents.

\textbf{A2A} (Agent2Agent, Google, April 2025; donated to Linux Foundation)~\citep{a2a2025}
is a \emph{horizontal} agent-to-agent protocol.
Any A2A server publishes an \textbf{AgentCard} at \texttt{/.well-known/agent-card.json}
declaring its skills, supported MIME types, transport bindings, and security schemes.
Agents communicate via JSON-RPC 2.0 (or gRPC / HTTP+JSON bindings) using an eight-state
\textbf{Task} lifecycle: \texttt{submitted}, \texttt{working}, \texttt{input\_required},
\texttt{auth\_required}, \texttt{completed}, \texttt{failed}, \texttt{canceled},
\texttt{rejected}.
Long-running tasks are tracked across connections via SSE streaming and webhook
push-notification configurations.
The A2A/MCP distinction is crisp: MCP is the ``USB-C for tool connectivity'' (vertical,
agent$\leftrightarrow$tool), while A2A is ``HTTP for agent collaboration'' (horizontal,
agent$\leftrightarrow$agent).
Production systems routinely combine both: A2A routes a task to the right specialist agent;
MCP gives that agent its context and tools.

\textbf{ACP} (Agent Communication Protocol, IBM/BeeAI, March 2025)~\citep{acp2025}
emphasizes REST-first, async-first enterprise messaging with multipart MIME payloads and
offline agent discovery for air-gapped deployments.
\textbf{ANP} (Agent Network Protocol)~\citep{anp2025} aims at decentralized open-internet
agent markets using W3C Decentralized Identifiers (DID) and JSON-LD agent descriptions.
\textbf{AGNTCY}~\citep{agntcy2025} (Cisco, now Linux Foundation) contributes the Open
Agent Schema Framework and a SLIM messaging layer; it is interoperable infrastructure
rather than a replacement for A2A or MCP.
All four are now converging under Linux Foundation governance.

Historically, \textbf{KQML}~\citep{kqml1994} and \textbf{FIPA-ACL}~\citep{fipaacl2002}
established the foundational concepts: performative-based communication (speech acts),
formal semantics grounded in agents' beliefs and intentions, and reusable multi-step
interaction protocols (contract net, query, propose).
FIPA-ACL's separation of communicative act from content language, and its contract-net
protocol for task delegation, are direct conceptual ancestors of A2A's task lifecycle.
However, FIPA's full BDI mental-state semantics proved impractical to implement and verify
in production systems, and both standards predate the LLM era.

None of these protocols, historical or contemporary, models the physical world.
An instrument is not a software agent: it is a stateful, safety-critical, exclusively
owned, physically embodied peer whose operations consume real reagents, take real time,
can injure people, and produce physically typed measurements with units and uncertainty.
LAP takes A2A's peer-to-peer, discovery-first, task-lifecycle spirit as its structural
skeleton and adds the four primitives the physical world demands (detailed in
\autoref{sec:protocol}).

\subsection{Agentic Protocols for Science}
\label{sec:related:science}

The closest prior work to LAP addresses the agent-to-instrument interface specifically in
scientific contexts; we compare each directly and state where LAP differs.

\citet{zhang2025mcp} propose using MCP as a natural-language interface for SDL
instruments, demonstrating that wrapping instrument APIs as MCP Tools enables LLM-driven
control with a shared protocol surface.
This confirms the value of a standard interface; the architecture is MCP-native (hub-and-spoke,
hub-mediated, agent-to-tool).
Because MCP Tool descriptions carry only JSON Schema and free-text, there is no
physical-quantity typing in parameters, no safety classification per capability, no
exclusive reservation primitive, and no calibration- or uncertainty-anchored result schema.

\citet{scp2025} introduce the \textbf{Science Context Protocol (SCP)}, the most
comprehensive existing science-domain protocol proposal.
Developed by the Shanghai Artificial Intelligence Laboratory and released in December 2025,
SCP extends MCP with a centralized SCP Hub that acts as a global registry, adds persistent
experiment objects with a five-phase lifecycle (Registration, Planning, Execution,
Monitoring, Archival), and provides a vendor-agnostic device-driver abstraction layer for
physical instruments.
SCP is currently deployed on the Intern-Discovery platform with over 1,600 integrated
tools spanning biology, physics, chemistry, and materials science.
We commend SCP as a serious and practical contribution.
The architectural difference from LAP is structural: SCP extends the \emph{MCP
agent-to-tool, hub-mediated} model upward, meaning all instrument access is proxied
through the Hub, capability negotiation is between the agent and the Hub rather than
directly with the instrument, and the Hub becomes a single point of trust and potential
failure for federated operation.
LAP follows the \emph{A2A peer-to-peer} line: each instrument is a first-class agent
that publishes its own signed \textbf{InstrumentCard}, and a Research Agent negotiates
directly with an Instrument Agent (or with a Lab Coordinator as an optional
orchestration layer that is never the sole path to the instrument).
Beyond architecture, LAP adds four capabilities absent from SCP:
\begin{enumerate}
  \item \textbf{Physical-quantity typing:} every parameter and result value carries a
        mandatory QUDT/UCUM unit annotation; the InstrumentCard declares physical limits
        (temperature range, maximum pressure, wavelength bounds) that are machine-checked
        before task submission (not present in SCP's tool description layer).
  \item \textbf{Exclusive reservation:} a first-class lease mechanism
        (\texttt{reservation.\allowbreak request} /\allowbreak{} \texttt{reservation.\allowbreak renew}
        /\allowbreak{} \texttt{reservation.\allowbreak release})
        prevents two agents from commanding the same physical instrument simultaneously, a
        safety property that cannot be expressed in MCP's stateless tool-call model.
  \item \textbf{Safety-fence handshake:} capabilities are classified S0--S3 by hazard
        level; S2/S3 operations require a \emph{bound operator-confirmation token}
        (JWS-signed by a human Safety Authority, cryptographically bound to the exact task
        and parameter hash) before the Instrument Agent will execute: a protocol-level
        interlock, not a UI-level hint.
  \item \textbf{Calibration- and uncertainty-anchored results:} every
        \texttt{lap:MeasurementResult} artifact carries a \texttt{calibrationRef},
        propagated measurement \texttt{uncertainty}, a full provenance manifest (parameter
        hash, instrument firmware, environmental conditions), and an instrument signature,
        making results FAIR, reproducible, and machine-verifiable in ways neither MCP
        tool outputs nor SCP experiment records currently mandate.
\end{enumerate}

\textbf{INTERSECT} (Interconnected Science Ecosystem, Oak Ridge National Laboratory)~\citep{intersect,intersect_sc25} is a deployed federated architecture for autonomous cross-facility scientific experimentation, integrating HPC resources, manufacturing user facilities, and, since 2025, LLM-based AI agents~\citep{intersect_sc25}.
INTERSECT operates as a microservices system: a natural-language chat assistant, a multi-agent orchestrator, programmable facility APIs (S3M), and a provenance-aware infrastructure (Flowcept) are composed into adaptive, explainable, reproducible workflows spanning ORNL's Manufacturing Demonstration Facility and the Oak Ridge Leadership Computing Facility.
INTERSECT is an \emph{institutional system and microservices architecture} with a running production deployment at national-laboratory scale; LAP is to INTERSECT what HTTP is to a specific server stack: a wire protocol that, if adopted, would let any INTERSECT facility API expose its instruments as first-class, reservation-aware, safety-fenced LAP agents discoverable by external research agents without INTERSECT-specific adapters.

\citet{vision2024nsls} describe \textbf{VISION}, a modular LLM-based assistant deployed at the 11-BM CMS beamline at NSLS-II (Brookhaven National Laboratory) that demonstrated the first voice-controlled beamtime, combining natural-language query with Bluesky-driven data acquisition.
VISION independently arrives at the ``confirm before actuate'' insight that LAP formalizes as the \texttt{intent.resolve} $\to$ \texttt{task.submit} two-step: natural language is resolved to a structured proposed action that a human or agent must confirm before hardware executes.

\citet{eac2025} frame the \textbf{Experiment-as-Code} (EaC) paradigm: treating the
laboratory as managed infrastructure analogous to Infrastructure-as-Code in cloud
computing, with declarative experiment specifications validated for safety before
actuation.
EaC motivates the same need LAP addresses: \textit{``laboratory instruments expose diverse
vendor APIs with inconsistent command interfaces, telemetry formats, and safety
semantics''}, and proposes declarative safety validation at the orchestration layer.
LAP is complementary: EaC defines the \emph{experiment specification language}; LAP
defines the \emph{agent-to-instrument wire protocol} that executes it.

Taken together, the closest prior science-agent work establishes that the field has
converged on the problem but remains split between hub-mediated MCP extensions (SCP,
\citet{zhang2025mcp}), deployed institutional systems (INTERSECT), and declarative
orchestration layers (EaC), while the ``confirm before actuate'' pattern has been
independently validated at the beamline level (VISION).
LAP occupies the orthogonal position: a peer-to-peer, instrument-as-agent \emph{wire protocol} whose capability-description layer is a profile of W3C WoT~TD~1.1 (not a novel contribution) and whose typed-data concepts build on AnIML/Allotrope conventions (equally established).
The new contribution is the \emph{combination} of primitives no prior protocol provides as a coherent agent-to-instrument layer: \textbf{(a)} a safety-fence handshake with operator-confirmation tokens cryptographically bound to exact task parameters, making hazardous-hardware autonomy auditable and bounded at the protocol level, and \textbf{(b)} a first-class reservation primitive that prevents concurrent physical actuation.
Together with physical-quantity typing (QUDT/UCUM mandatory) and calibration-and-uncertainty-anchored measurement results, these four primitives enable autonomous, safe, and auditable operation of physical laboratory instruments across institutional boundaries.

% ============================================================
\section{Design Overview}
\label{sec:overview}

\subsection{The third edge}
\label{sec:thirdedge}
\begin{figure}[t]
\centering
\adjustbox{max width=\linewidth}{%
\begin{tikzpicture}[node distance=22mm]
  \node[role,fill=laplight] (agent) at (0,2.5) {Reasoning\\Agent};
  \node[role,fill=laplight] (peer)  at (-4.2,-1.6) {Peer\\Agent};
  \node[box,fill=laplight2,minimum width=20mm] (tool) at (4.2,-1.6) {Tool /\\Service};
  \node[box,draw=lapamber!80,fill=lapamber!12,minimum width=22mm] (instr) at (0,-1.6)
       {\textbf{Physical}\\\textbf{Instrument}};
  % edges
  \draw[{Stealth[length=2.5mm]}-{Stealth[length=2.5mm]},very thick,lapteal]
     (agent) -- (peer) node[midway,above,sloped,font=\scriptsize\bfseries,lapteal]{A2A: agent$\leftrightarrow$agent};
  \draw[{Stealth[length=2.5mm]}-{Stealth[length=2.5mm]},very thick,lapblue]
     (agent) -- (tool) node[midway,above,sloped,font=\scriptsize\bfseries,lapblue]{MCP: agent$\leftrightarrow$tool};
  \draw[{Stealth[length=2.5mm]}-{Stealth[length=2.5mm]},very thick,lapamber]
     (agent) -- (instr) node[midway,fill=white,text=lapamber,inner sep=2pt,font=\scriptsize\bfseries]{LAP: agent$\leftrightarrow$instrument};
  \node[font=\scriptsize,lapgray,align=center] at (0,-3.0)
     {\emph{stateful $\cdot$ exclusive $\cdot$ safety-critical $\cdot$ physically typed}};
\end{tikzpicture}%
}
\caption{The three edges of an agentic scientific ecosystem. MCP standardizes the
agent-to-tool edge and A2A the agent-to-agent edge; LAP completes the triangle with the
agent-to-instrument edge, whose distinguishing properties (annotated below the
instrument node) drive LAP's four physical-world primitives.}
\label{fig:triangle}
\end{figure}
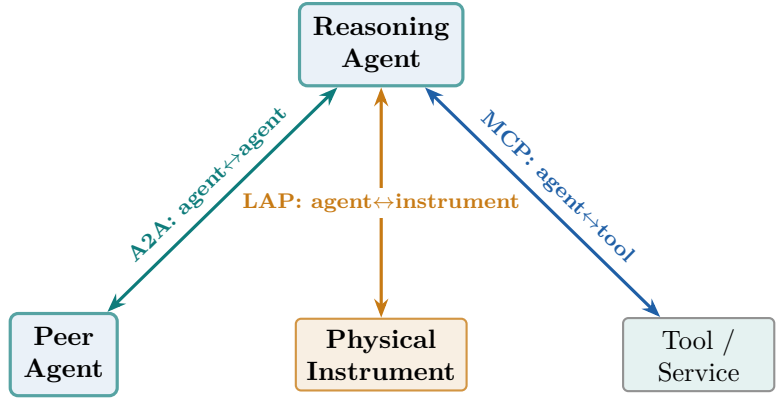

We organize an agentic scientific ecosystem around three communication edges
(\autoref{fig:triangle}). The \emph{agent-to-tool} edge, standardized by MCP, lets a
single reasoning agent call functions, read resources, and invoke prompts on services it
controls. The \emph{agent-to-agent} edge, standardized by A2A, lets autonomous agents
treat one another as peers, advertising capabilities through an AgentCard, delegating
work as tasks, and exchanging messages and artifacts. The \emph{agent-to-instrument}
edge, which LAP defines, connects a reasoning agent to a physical instrument. It is not
reducible to either neighbor: a tool call is stateless and instantaneous, a peer agent
is itself a reasoner, but an instrument is a stateful, exclusively owned, safety-critical
physical resource whose operations are slow, consequential, and measured in physical
units.

From these properties we derive four primitives that any agent-to-instrument protocol
must provide, and that organize the rest of this report:
\begin{description}[leftmargin=1.6em,itemsep=2pt]
\item[\textnormal{(P1) Capability and physical-limit description.}] An agent cannot be
  preprogrammed for every instrument; it must discover, at run time, what an instrument
  can do, within what physical bounds, with what safety implications, and how to express
  a goal to it. LAP answers this with the \emph{InstrumentCard}.
\item[\textnormal{(P2) Exclusive reservation.}] A physical instrument can serve one
  experiment at a time; two agents cannot interleave operations on a single
  diffractometer. LAP makes \emph{reservation} a first-class, leased primitive.
\item[\textnormal{(P3) Safety negotiation.}] Some operations are hazardous or
  irreversible. Autonomous authority over them must be explicit and bounded. LAP defines
  a \emph{safety-fence handshake} with task-bound operator-confirmation tokens.
\item[\textnormal{(P4) Physically honest results.}] A number without a unit, a
  calibration, and an uncertainty is not a measurement. LAP's \emph{MeasurementResult}
  makes all four mandatory.
\end{description}

\subsection{Roles}
\label{sec:roles}
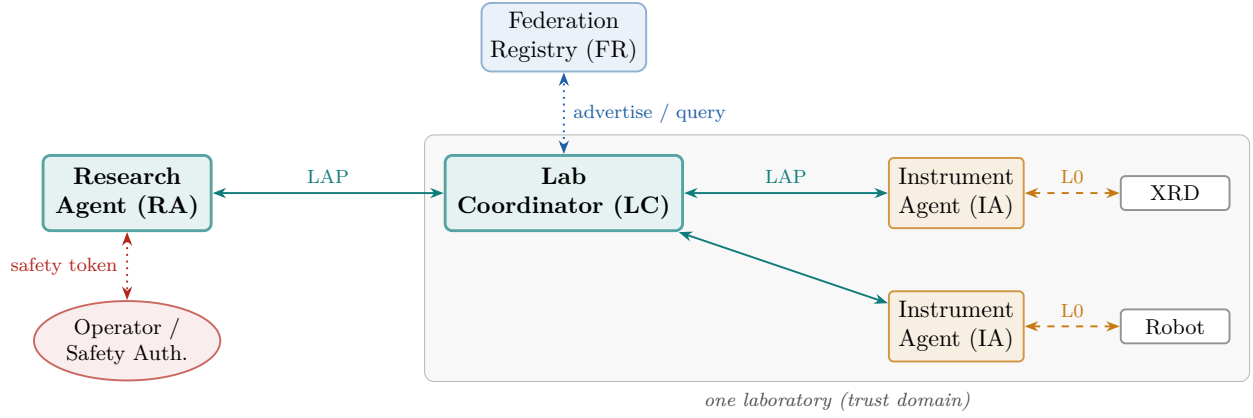
\begin{figure}[t]
\centering
\adjustbox{max width=\linewidth}{%
\begin{tikzpicture}[node distance=14mm and 20mm]
  \node[role] (ra) {Research\\Agent (RA)};
  \node[role,right=34mm of ra] (lc) {Lab\\Coordinator (LC)};
  \node[box,draw=lapamber!80,fill=lapamber!12,right=30mm of lc] (ia1) {Instrument\\Agent (IA)};
  \node[box,below=9mm of ia1,draw=lapamber!80,fill=lapamber!12] (ia2) {Instrument\\Agent (IA)};
  \node[box,fill=white,right=14mm of ia1,minimum width=16mm] (dev1) {\footnotesize XRD};
  \node[box,fill=white,right=14mm of ia2,minimum width=16mm] (dev2) {\footnotesize Robot};
  \node[draw=lapblue!50,thick,fill=laplight,rounded corners,above=12mm of lc,align=center,font=\small]
       (fr) {Federation\\Registry (FR)};
  \node[term,draw=lapred!70,fill=lapred!8,below=10mm of ra,font=\footnotesize] (ho) {Operator /\\Safety Auth.};

  \draw[{Stealth[length=2mm]}-{Stealth[length=2mm]},thick,lapteal] (ra) -- (lc)
       node[midway,above,font=\scriptsize]{LAP};
  \draw[{Stealth[length=2mm]}-{Stealth[length=2mm]},thick,lapteal] (lc) -- (ia1)
       node[midway,above,font=\scriptsize]{LAP};
  \draw[{Stealth[length=2mm]}-{Stealth[length=2mm]},thick,lapteal] (lc) -- (ia2);
  \draw[{Stealth[length=2mm]}-{Stealth[length=2mm]},thick,lapamber,dashed] (ia1) -- (dev1)
       node[midway,above,font=\scriptsize]{L0};
  \draw[{Stealth[length=2mm]}-{Stealth[length=2mm]},thick,lapamber,dashed] (ia2) -- (dev2)
       node[midway,above,font=\scriptsize]{L0};
  \draw[{Stealth[length=2mm]}-{Stealth[length=2mm]},thick,lapblue,dotted] (lc) -- (fr)
       node[midway,right,font=\scriptsize]{advertise / query};
  \draw[{Stealth[length=2mm]}-{Stealth[length=2mm]},thick,lapred,dotted] (ra) -- (ho)
       node[midway,left,font=\scriptsize]{safety token};
  \begin{scope}[on background layer]
    \node[draw=lapgray!40,rounded corners,fill=lapgray!4,fit=(lc)(ia1)(ia2)(dev1)(dev2),
          inner sep=8pt,label={[font=\scriptsize\itshape,lapgray]below:one laboratory (trust domain)}] {};
  \end{scope}
\end{tikzpicture}%
}
\caption{LAP role topology. A Research Agent reaches Instrument Agents directly or
through an optional Lab Coordinator; each Instrument Agent encapsulates one physical
device through a device-abstraction layer (L0) that the agent never sees. A laboratory
is a trust domain that advertises capabilities to a Federation Registry; the Operator /
Safety Authority is the trust anchor for safety-fenced operations.}
\label{fig:topology}
\end{figure}

LAP defines five roles (\autoref{fig:topology}).
\begin{itemize}[leftmargin=1.4em,itemsep=2pt]
\item The \textbf{Research Agent (RA)} is the autonomous scientist. It forms hypotheses,
  plans campaigns, decomposes them into instrument tasks, interprets results, and drives
  the closed loop. It speaks LAP as a client.
\item The \textbf{Instrument Agent (IA)} wraps exactly one physical instrument (or one
  logical cluster). It translates LAP tasks into vendor-SDK, SCPI, serial, OPC-UA, or
  SiLA\,2 calls below it, and it owns the instrument's state, safety interlocks,
  calibration, and reservation table. It publishes an InstrumentCard and speaks LAP as a
  server.
\item The \textbf{Lab Coordinator (LC)} is a per-laboratory broker, optional for a single
  instrument but \emph{required} for multi-instrument workflows that carry cross-instrument
  hazards. It aggregates InstrumentCards, arbitrates contended reservations, enforces
  lab-wide safety policy, and brokers sample chain-of-custody between instruments. Because
  it is the only role that sees a workflow as a whole, it is also the enforcement point for
  \emph{workflow-level} (class~(C)) safety: before routing a sample or granting a sequence
  of reservations it MUST evaluate cross-instrument safety predicates---e.g.\ checking the
  \texttt{sampleCondition} flags carried on a \texttt{MeasurementResult} against the next
  instrument's \texttt{physicalLimits}, and enforcing cumulative cost or exposure
  budgets---that no single Instrument Agent can see. It presents the laboratory as a single
  federable endpoint, and is a LAP server to RAs and a LAP client to IAs.
\item The \textbf{Federation Registry (FR)} is an optional cross-laboratory directory. It
  indexes instruments by capability, location, availability, and access policy, and
  returns signed pointers; it never proxies control.
\item The \textbf{Human Operator / Safety Authority (HO)} is not an agent but a trust
  anchor: it issues or relays the operator-confirmation tokens that gate safety-fenced
  operations and can assert emergency stop.
\end{itemize}
The minimal valid topology is a single RA talking directly to a single IA; the LC and FR
are layered in as a laboratory grows and federates. This mirrors how the web grew from
isolated servers into a federated network. Single-instrument operations (classes (A) and
(B) safety) are fully protected in this minimal topology; workflow-level (class (C)) safety,
by contrast, presupposes a Lab Coordinator, since cross-instrument hazards are invisible to
any single Instrument Agent.

\Needspace*{10\baselineskip}
\subsection{A layered architecture}
\label{sec:layers}
\begin{figure}[t]
\centering
\adjustbox{max width=\linewidth}{%
\begin{tikzpicture}[node distance=2mm]
  \node[layer,fill=lapgray!8,draw=lapgray!50,text width=198mm,align=left] (l5)
    {\textbf{L5 \ Campaign / Closed-Loop} \;\textemdash\; experiment design, active learning, iteration \hfill\emph{(RA-internal; out of scope)}};
  \node[layer,below=of l5,text width=198mm,align=left] (l4)
    {\textbf{L4 \ Orchestration} \;\textemdash\; multi-instrument workflows, sample routing, scheduling, reservations \hfill\emph{(LC)}};
  \node[layer,below=of l4,text width=198mm,align=left,fill=laplight2,draw=lapteal!50] (l3)
    {\textbf{L3 \ Semantic} \;\textemdash\; InstrumentCard, capability ontology, QUDT/UCUM typing, MeasurementResult};
  \node[layer,below=of l3,text width=198mm,align=left,fill=laplight2,draw=lapteal!50] (l2)
    {\textbf{L2 \ Interaction} \;\textemdash\; LAP methods, Task lifecycle, safety-fence handshake, streaming, events};
  \node[layer,below=of l2,text width=198mm,align=left,fill=laplight2,draw=lapteal!50] (l1)
    {\textbf{L1 \ Identity \& Transport} \;\textemdash\; JSON-RPC\,2.0 / HTTPS\,+\,SSE\,+\,webhooks; mTLS / OAuth\,2.1 / DIDs; \texttt{/.well-known}};
  \node[layer,below=of l1,text width=198mm,align=left,fill=lapgray!8,draw=lapgray!50] (l0)
    {\textbf{L0 \ Device Abstraction} \;\textemdash\; vendor SDK / SCPI / serial / GPIB / OPC-UA / SiLA\,2 adapter \hfill\emph{(inside IA; out of scope)}};
  \draw[decorate,decoration={brace,amplitude=5pt},lapteal!70,thick]
       ($(l3.east)+(3mm,2mm)$) -- ($(l1.east)+(3mm,-2mm)$);
  \node[rotate=-90,anchor=center,font=\scriptsize\bfseries,lapteal] at ($(l3.east)!0.5!(l1.east)+(12mm,0)$) {LAP normative};
\end{tikzpicture}%
}
\caption{The LAP six-layer architecture. LAP normatively specifies L1--L3 and the L4
orchestration messages (teal). L0 (device abstraction, inside the Instrument Agent) and
L5 (the science, inside the Research Agent) are intentionally out of scope: encapsulating
L0 is the protocol's purpose, and serving (without dictating) L5 keeps the protocol
science-agnostic.}
\label{fig:stack}
\end{figure}
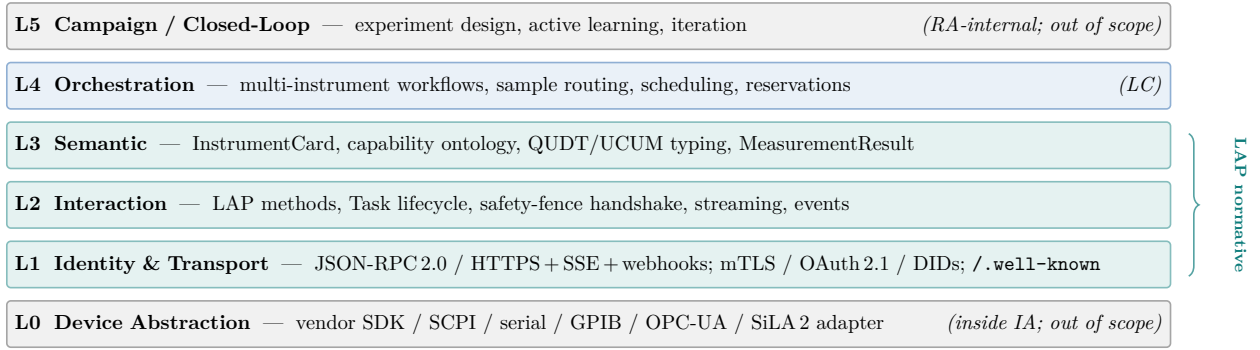

LAP is organized into six layers (\autoref{fig:stack}). It normatively defines the
identity/transport layer (L1), the interaction layer (L2), and the semantic layer (L3),
together with the orchestration messages of L4. The device-abstraction layer (L0) and
the campaign/closed-loop layer (L5) are deliberately \emph{out of scope}: L0 is the
Instrument Agent's private concern (encapsulating it is the entire point, so that a
Research Agent never sees a vendor SDK) and L5 is the science, which LAP serves with
hooks but does not dictate. This boundary is what lets a single InstrumentCard expose a
SiLA\,2 instrument, a SCPI oscilloscope, and a custom robot through one uniform agentic
interface.

\subsection{Identity and naming}
\label{sec:naming}
Every LAP entity has a stable URI under a laboratory authority:
an instrument is \nolinkurl{lap://<lab>/instruments/xrd-bruker-d8-01}, a capability
is a path beneath it, a sample is \nolinkurl{lap://<lab>/samples/2026-0531-A7}, and
a task is \nolinkurl{lap://<lab>/tasks/UUID}. Each Instrument Agent, Research
Agent, and Lab Coordinator carries a decentralized identifier
(\nolinkurl{did:web:<lab>:instruments:xrd-01}) or an equivalent X.509 identity, and
every InstrumentCard is signed by its laboratory authority so that a remote Research
Agent can verify provenance before trusting a capability advertisement. This naming and
identity scheme is the substrate on which the federation of \autoref{sec:federation} is
built.

% ============================================================
\section{Protocol Specification}
\label{sec:protocol}

This section specifies LAP's semantic and interaction layers: the InstrumentCard (P1),
the method set, the reservation primitive (P2), the task and safety state machines (P3),
the MeasurementResult (P4), and the error model. Throughout, we reuse JSON-RPC\,2.0 as
the message envelope, JSON Schema 2020-12 for parameter typing, JSON-LD with a LAP
context for semantic annotation, and UCUM codes with QUDT \texttt{quantityKind} URIs for
physical quantities.

\subsection{The InstrumentCard (P1)}
\label{sec:card}
The InstrumentCard is LAP's central artifact: the instrument analog of an A2A AgentCard,
realized as a profile of the W3C Web of Things Thing Description. It is served at
\nolinkurl{https://<host>/.well-known/instrument-card.json} and is signed by the
laboratory authority. Beyond the identity, transport, and security fields it shares with
an AgentCard, it adds the instrument-specific structure that an autonomous agent needs:
an \texttt{instrumentClass} drawn from a capability ontology; a list of
\texttt{capabilities}, each carrying physically typed input/output schemas, a
\texttt{safetyClass}, a \texttt{reversible} flag, an \texttt{estimatedDuration},
\texttt{physicalLimits} and \texttt{interlocks}, and natural-language \texttt{intentTags}
for grounding; continuous telemetry \texttt{streams}; and a \texttt{calibration} block.
\autoref{lst:card} shows the skeleton of a card for a powder X-ray diffractometer:
its identity, one physically typed capability with its safety class and intent tags, and the
signature envelope. The complete card, with full input schemas, telemetry streams, and
calibration block, is given in \autoref{lst:card-full} (\autoref{app:schemas}).

\noindent\begin{minipage}{\linewidth}
\begin{lstlisting}[caption={Skeleton of an InstrumentCard for a powder XRD instrument:
   identity and instrument class, one physically typed capability (\texttt{powder-scan})
   with its safety class, intent tags, and physical limits, and the JWS signature
   envelope. The complete card is given in \autoref{lst:card-full}.},label={lst:card}]
{
  "@type": "lap:InstrumentCard",
  "id": "lap://<lab>/instruments/xrd-bruker-d8-01",
  "instrumentClass": "xrd:PowderDiffractometer",
  "lapProfile": {"streaming": true, "reservation": true,
                 "safetyFence": true, "federation": true,
                 "intentResolve": true},
  "capabilities": [{
    "id": "powder-scan",
    "intentTags": ["measure crystal structure", "phase identification"],
    "inputSchema": {"twoThetaStart": {"$ref": "lap:Quantity",
                    "unit": "deg", "minimum": 5, "maximum": 120}},
    "outputSchema": {"$ref": "lap:MeasurementResult"},
    "safetyClass": "S1", "reversible": true,
    "physicalLimits": {"maxPower": "2.2 kW",
                       "interlocks": ["shutter", "doorClosed"]}
  }],
  "calibration": {"calibrationRef": "lap://<lab>/cal/xrd01/2026-05-20"},
  "signatures": [{"alg": "ES256", "by": "did:web:<lab>", "jws": "..."}]
}
\end{lstlisting}
\end{minipage}

\subsection{Method set}
\label{sec:methods}
All LAP interactions are JSON-RPC\,2.0 requests; streaming methods upgrade to
Server-Sent Events, and long-running tasks may additionally register a webhook.
\autoref{tab:methods} lists the method set, grouped by the primitive it serves.

\begin{table}[t]
\centering\small
\begin{tabular}{@{}lll@{}}
\toprule
\textbf{Method} & \textbf{Direction} & \textbf{Purpose}\\
\midrule
\texttt{instrument.describe} & RA$\to$IA & Fetch/refresh the InstrumentCard.\\
\texttt{instrument.getState} & RA$\to$IA & Operational + safety + calibration snapshot.\\
\addlinespace
\texttt{reservation.request} & RA$\to$IA/LC & Acquire an exclusive (or shared-read) lease.\\
\texttt{reservation.renew}   & RA$\to$IA & Extend a lease (heartbeat) before expiry.\\
\texttt{reservation.release} & RA$\to$IA & Free a held lease.\\
\addlinespace
\texttt{task.submit}        & RA$\to$IA & Invoke a capability (typed params or proposal ref).\\
\texttt{task.get}           & RA$\to$IA & Poll task status and artifacts.\\
\texttt{task.stream}        & RA$\to$IA (SSE) & Subscribe to live status, stream frames, events.\\
\texttt{task.cancel}        & RA$\to$IA & Cooperative graceful cancellation.\\
\texttt{task.subscribePush} & RA$\to$IA & Register a webhook for asynchronous completion.\\
\addlinespace
\texttt{safety.requestAuthorization} & IA$\to$RA & Demand an operator-confirmation token.\\
\texttt{safety.provideToken}         & RA$\to$IA & Return a signed operator-confirmation token.\\
\texttt{safety.emergencyStop}        & RA/HO$\to$IA & Immediate halt; bypasses the queue.\\
\addlinespace
\texttt{intent.resolve} & RA$\to$IA & Map a natural-language goal to a proposed \texttt{task.submit}.\\
\addlinespace
\texttt{registry.query}     & RA$\to$FR & Find instruments by capability/location/availability.\\
\texttt{registry.advertise} & LC$\to$FR & Publish/refresh a laboratory's instruments.\\
\bottomrule
\end{tabular}
\caption{The LAP method set. Every state-changing physical action is expressed as a
\texttt{Task}, making it auditable, cancellable, and streamable. Natural language enters
only through \texttt{intent.resolve}, which returns a \emph{proposed} structured call
that the agent must confirm via \texttt{task.submit}; natural language never drives
hardware directly.}
\label{tab:methods}
\end{table}

\paragraph{Reservation (P2).} Because an instrument serves one experiment at a time,
\texttt{reservation.request} returns a \texttt{reservationToken} and a time-boxed lease;
\texttt{task.submit} on an exclusive instrument requires a valid token, and the lease
must be kept alive with \texttt{reservation.renew} or it expires and frees the
instrument. This converts the implicit, error-prone mutual exclusion that orchestrators
hand-code today into an explicit protocol primitive with conflict and expiry semantics
(error codes in \autoref{sec:errors}).

\paragraph{Natural-language grounding at the edge.} A central design decision is that
natural language is confined to \texttt{intent.resolve}, which takes a free-text goal
(``acquire a fine-step powder pattern from 10 to 80 degrees'') and returns a
\emph{proposed} structured \texttt{task.submit} payload (a chosen capability with filled,
typed, limit-checked parameters) that the Research Agent must explicitly confirm.
Hardware is therefore only ever actuated through a validated structured call, never
directly from model output. This is primarily a \emph{reliability} decision: it confines
the stochasticity of a language model to a proposal that is inspected and replayable
(we develop this in \autoref{sec:discussion}). It is deliberately \emph{not} the protocol's
safety guarantee. \texttt{intent.resolve} is best-effort and optional, and a proposal
carries no privileged trust; the deterministic \texttt{task.submit} validation gate that
follows it---schema and \texttt{physicalLimits} checking, and, for S2/S3 capabilities, the
operator-token handshake---is the \emph{precondition} on which the safety mechanisms of
\autoref{sec:safety} operate, and it runs identically whether the parameters came from
\texttt{intent.resolve}, from the agent's own grounding against the InstrumentCard, or
from a human.

\subsection{The \texttt{intent.resolve} method}
\label{sec:intent}
\texttt{intent.resolve} is the one method that accepts free text: it maps a natural-language
goal to a \emph{proposed} \texttt{task.submit}---a chosen capability with filled, typed,
limit-checked parameters---that the Research Agent inspects and confirms. It is optional; an
IA advertises support with the \texttt{intentResolve} flag in its \texttt{lapProfile}, and an
IA that does not implement it (a deterministic or legacy instrument) simply returns JSON-RPC
\texttt{-32601 (Method not found)}, leaving the agent to build the \texttt{task.submit} from
the card's \texttt{intentTags} and \texttt{inputSchema} directly. The proposal carries no
privileged trust: because the deterministic \texttt{task.submit} gate (\autoref{sec:safety})
re-validates the parameters either way, \texttt{intent.resolve} is a convenience, not a
safety mechanism. A clean resolution returns the proposal directly:

\noindent\begin{minipage}{\linewidth}
\begin{lstlisting}[caption={A clean \texttt{intent.resolve}: the free-text goal (top) and the
   proposed \texttt{task.submit} the agent confirms (bottom). The IA surfaces any value it
   had to fill in as an \texttt{assumption}.},label={lst:intent-resp}]
{"goal": "fine-step powder pattern from 10 to 80 degrees",
 "capabilityHint": "powder-scan"}

{"status": "resolved",
 "resolutionId": "res-7f3a",
 "proposal": {
   "capability": "powder-scan",
   "params": {
     "twoThetaStart": {"value": 10,   "unit": "deg"},
     "twoThetaEnd":   {"value": 80,   "unit": "deg"},
     "stepSize":      {"value": 0.01, "unit": "deg"}
   },
   "assumptions": ["'fine' step taken as 0.01 deg (capability default)"]
 }
}
\end{lstlisting}
\end{minipage}

To act on a resolution the Research Agent does not re-author the structured call: it commits
\emph{by reference}, submitting \texttt{task.submit} with the proposal's
\texttt{resolutionId} and a \texttt{confirm} flag rather than a hand-built parameter block.
The commit is still the agent's own accountable act---signed with its DID and bound to that
exact, immutable proposal---so the provenance records that the Research Agent, not the IA,
decided to actuate, and the IA executes precisely the proposal that was shown. An agent that
wants to change a value, or a deterministic client that never called \texttt{intent.resolve},
may instead submit a full typed \texttt{task.submit} directly; both forms reach the same
validation gate. The common LLM-agent path is thus natural language end to end---goal in,
proposal reviewed, one confirmation---with the typed task existing as the signed record
beneath it rather than as something the agent must build.

When the goal is under-determined, the IA does not guess silently: it returns the slots it
could not fill as \texttt{clarifications}, with the options it can offer, and the agent
answers (re-calling with the same \texttt{resolutionId}) or accepts a default before
submitting:

\noindent\begin{minipage}{\linewidth}
\begin{lstlisting}[caption={An under-determined goal. Rather than guess, the IA returns the
   missing parameters as \texttt{clarifications} for the agent---or a human---to
   resolve.},label={lst:intent-ambig}]
{"goal": "scan this sample"}

{"status": "ambiguous",
 "resolutionId": "res-7f3a",
 "proposal": {"capability": "powder-scan", "params": {}},
 "clarifications": [
   {"param": "twoThetaStart", "question": "start angle?", "unit": "deg"},
   {"param": "twoThetaEnd",   "question": "end angle?",   "unit": "deg"},
   {"param": "stepSize",      "question": "fine (0.01) or coarse (0.05)?",
    "options": [{"value": 0.01, "unit": "deg"}, {"value": 0.05, "unit": "deg"}]}
 ]
}
\end{lstlisting}
\end{minipage}

If the goal maps to no capability the instrument has, the call fails with
\texttt{IntentUnresolved} (\autoref{sec:errors}), whose \texttt{reason} lets the agent
re-plan against a different instrument rather than retry blindly. \texttt{intent.resolve}
creates no task and changes no instrument state.

\subsection{Reservation and resource locking (P2)}
\label{sec:reservation}
Because a physical instrument can serve only one experiment at a time, LAP makes mutual
exclusion an explicit, leased primitive rather than leaving it to convention. A
\texttt{reservation.request} carries a requested \texttt{mode}, a \texttt{duration}, and
the target resource (an instrument and, optionally, a sample), and an optional
\texttt{budgetCap} (a currency and an amount) that bounds the cumulative
\texttt{operationalCost} (declared per capability in the InstrumentCard) of all tasks
submitted under the lease; on success it returns a \texttt{reservationToken}:

\noindent\begin{minipage}{\linewidth}
\begin{lstlisting}
{
  "id": "<uuid>",
  "resource": "<instrument-uri>",
  "holder": "<agent-did>",
  "mode": "exclusive | shared-read",
  "grantedAt": "<ts>",
  "expiresAt": "<ts>",
  "epoch": "<int>"
}
\end{lstlisting}
\end{minipage}

LAP defines two modes. An \textbf{exclusive} lease confers sole right to submit
state-changing tasks; while it is held, \texttt{reservation.request}s from other agents
fail with \texttt{ReservationConflict} (\autoref{sec:errors}), whose error data returns the
current \texttt{holder} and \texttt{expiresAt} so the requester can queue or back off. A
\textbf{shared-read} lease permits multiple concurrent holders to call read-only methods
(\texttt{instrument.getState}, telemetry \texttt{streams}) but no state-changing
\texttt{task.submit}; an exclusive request is refused while any shared-read lease is
active, and vice versa. Leases are time-boxed: the holder must call
\texttt{reservation.renew} before \texttt{expiresAt} or the lease lapses and the instrument
is freed, which prevents a crashed agent from deadlocking an instrument indefinitely. The
monotonic \texttt{epoch} is incremented on every grant so that a task submitted under a
lapsed-and-reissued lease is detected and rejected with \texttt{LeaseExpired}. Two
operations bypass reservation unconditionally: \texttt{instrument.describe} (discovery is
always permitted) and the S0 fast path: \texttt{safety.emergencyStop} bypasses both the
lease and the credential scope. Any authenticated party that holds, or has held within the
current \texttt{epoch}, a valid lease or scoped credential on the instrument may issue it,
and a conforming IA MUST honor it regardless of whether ``emergency stop'' was explicitly
enumerated in that credential's scope, because halting a runaway instrument must never be
blocked by a lock or by an under-scoped grant (this is the one method whose authorization
the IA does not scope-check). In federation, this ensures that a remote Research Agent
driving an instrument can always stop it even if its time-boxed credential was issued for a
single capability. A Lab Coordinator, when present, is the reservation
authority for its instruments and may additionally enforce lab-wide policies such as
fair-share quotas or priority pre-emption (a pre-empted holder is notified and its running
task transitions to \texttt{canceled}). When a lease carries a \texttt{budgetCap}, the IA
(or the LC, when it is the reservation authority) MUST refuse any \texttt{task.submit} that
would drive the lease's cumulative \texttt{operationalCost} past the cap, returning
\texttt{BudgetExceeded} (\autoref{sec:errors}); this bounds the class~(C) failure mode of an
autonomous loop consuming unbounded beam-time, reagents, or billable cost while each
individual step remains within physical limits and safety class~S1.

\subsection{Task lifecycle (state machine)}
\label{sec:tasksm}
\begin{figure}[t]
\centering
\adjustbox{max width=\linewidth}{%
\begin{tikzpicture}[node distance=12mm and 16mm]
  \node[state] (sub) {submitted};
  \node[state,right=of sub] (queue) {queued};
  \node[state,right=of queue] (run) {running};
  \node[sstate,above=14mm of run] (hold) {safety-\\hold};
  \node[sstate,below left=14mm and 9mm of run] (fault) {paused-\\fault};
  \node[sstate,above left=19mm and 9mm of queue] (swait) {sample-\\wait};
  \node[term,right=18mm of run] (done) {completed};
  \node[term,below=20mm of done] (failed) {failed};
  \node[term,below=9mm of sub] (rej) {rejected};

  \draw[flow] (sub) -- (queue) node[midway,above]{valid};
  \draw[flow] (sub) -- (rej) node[midway,left]{schema /};
  \node[font=\scriptsize,lapgray] at ($(sub)!0.5!(rej)+(11mm,0)$){limit fail};
  \draw[flow] (queue) -- (run) node[midway,above]{lease ok};
  \draw[flow] (swait) -- (queue) node[midway,above,sloped,font=\scriptsize]{sample present};
  \draw[flow] (queue) to[bend left=12] (swait);
  \draw[flow] (run) -- (done) node[midway,above]{result};
  \draw[flow,lapamber] (run) -- (hold) node[midway,right,align=left,font=\scriptsize]{S2/S3,\\no token};
  \draw[flow,lapamber] (hold) to[bend right=35] node[midway,left,font=\scriptsize]{token ok} (run);
  \draw[flow,lapamber] (hold) -| node[pos=0.28,above,font=\scriptsize]{deny/timeout} ($(done.east)+(6mm,0)$) -- (failed.east);
  \draw[flow,lapred] (run) -- (fault) node[midway,above,sloped,font=\scriptsize]{interlock};
  \draw[flow] (fault) to[bend right=18] node[midway,below,font=\scriptsize]{recover} (run);
  \draw[flow] (fault) -- (failed) node[midway,left,font=\scriptsize]{unrecoverable};
  \draw[flow,lapred,dashed] (run) to[bend left=10] node[midway,right,font=\scriptsize]{e-stop} (failed);
\end{tikzpicture}%
}
\caption{The LAP task lifecycle. Blue states and \texttt{completed}/\texttt{failed}/%
\texttt{rejected} follow A2A; the amber/red physical states (\texttt{safety-hold},
\texttt{paused-fault}, \texttt{sample-wait}) are LAP additions that capture safety
negotiation, interlock faults, and physical-sample dependencies. Emergency stop drives
any running task to \texttt{failed}.}
\label{fig:taskfsm}
\end{figure}
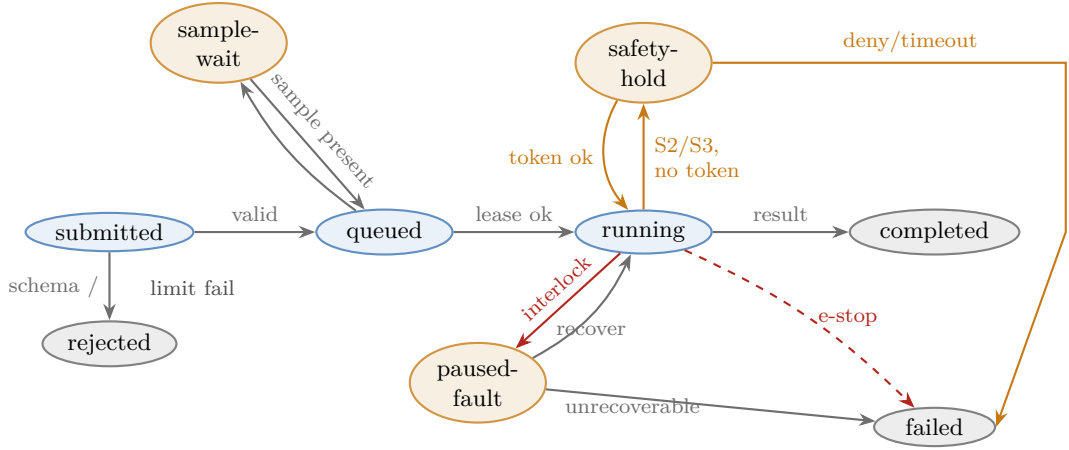

A LAP task extends the eight A2A task states with three states that only a physical
process can occupy (\autoref{fig:taskfsm}). After \texttt{submitted}, parameters are
validated against the capability schema and \texttt{physicalLimits}, and the IA MUST
additionally reject the task if the instrument's calibration has lapsed
(\texttt{calibration.validUntil} in the past), since a task run under expired calibration
yields a result that is physically valid but scientifically untrustworthy (class~(C)
integrity); any such failure yields \texttt{rejected} (\texttt{CalibrationExpired} for the
calibration case, \autoref{sec:errors}). A valid task moves through \texttt{queued} to \texttt{running}. Three
physical states branch off the normal path: \texttt{safety-hold}, entered when a
capability of class S2/S3 lacks a valid operator token and an authorization handshake is
in progress (\autoref{sec:safety}); \texttt{paused-fault}, entered when a hardware
interlock trips, from which the task may auto-recover to \texttt{running} or terminate as
\texttt{failed}; and \texttt{sample-wait}, entered when the task is blocked awaiting
physical delivery of a sample via chain-of-custody from another Instrument Agent. The
A2A interrupt states \texttt{input-required} and \texttt{auth-required} are retained. A
successful task terminates as \texttt{completed} and carries one or more
\texttt{MeasurementResult} artifacts; \texttt{safety.emergencyStop} drives any
non-terminal state to \texttt{failed} with an \texttt{eStop} flag.

\subsection{The safety fence (P3)}
\label{sec:safety}
LAP separates three kinds of safety and places each at the layer that can enforce it.
\textbf{(A) Device-physics safety}---never exceeding an instrument's physical envelope
(torque, voltage, temperature, pressure), respecting hardware interlocks, and aborting on
fault---is local to one instrument and instantaneous. LAP delegates it \emph{entirely} to
the Instrument Agent: the protocol imposes the normative requirement that an IA MUST reject
or abort any task whose parameters fall outside its declared \texttt{physicalLimits} or
whose interlocks are not satisfied (\autoref{sec:errors}), but treats the enforcement
mechanism as a black box, since the instrument vendor is best placed to implement it below
Layer~0. \textbf{(B) Authorization and accountability safety}---obtaining and recording a
human authority's consent for a hazardous, irreversible, or costly action---crosses the
human--agent trust boundary and must be uniform and verifiable across heterogeneous
vendors; LAP therefore standardizes it at the protocol level as the safety fence specified
in the remainder of this section. \textbf{(C) Workflow- and integrity-level safety}---hazards
that no single IA can see, such as a sequence of individually-safe steps that is unsafe in
combination, a ``physically valid but scientifically invalid'' result taken under expired
calibration, or unbounded resource or cost consumption across an autonomous campaign---is
the concern of the orchestration layer (the Lab Coordinator, \autoref{sec:roles}) together
with a few protocol-level declarations (\texttt{sampleCondition}, mandatory
calibration-expiry rejection, and \texttt{budgetCap}). The remainder of this section
specifies (B); (A) and (C) are referenced where the protocol carries an obligation or a
declaration that supports them.

Every capability declares a \texttt{safetyClass}: \textbf{S0} for emergency/meta
operations (e-stop, abort), always permitted and reachable on a fast path; \textbf{S1}
for routine, reversible, non-hazardous operations, requiring no fence; \textbf{S2} for
operations that consume an irreplaceable sample or are costly or hard to reverse,
requiring a fresh operator-confirmation token or a standing policy grant; and
\textbf{S3} for hazardous operations (lasers, high voltage, pyrophorics, high pressure,
radiation, or robotics acting near humans) requiring an operator-confirmation token
signed by a Safety Authority \emph{and} a positive interlock attestation in
\texttt{instrument.getState}, never auto-grantable by the Research Agent alone.

The operator-confirmation token is a JWS whose claims bind the approval to one exact
action:

\noindent\begin{minipage}{\linewidth}
\begin{lstlisting}
{
  "jti": "<nonce>",
  "sub": "<task-uri>",
  "instr": "<instrument-uri>",
  "cap": "<capability-id>",
  "paramsHash": "<sha256>",
  "authority": "<did>",
  "iat": "<ts>",
  "exp": "<ts>",
  "scope": ["..."]
}
\end{lstlisting}
\end{minipage}

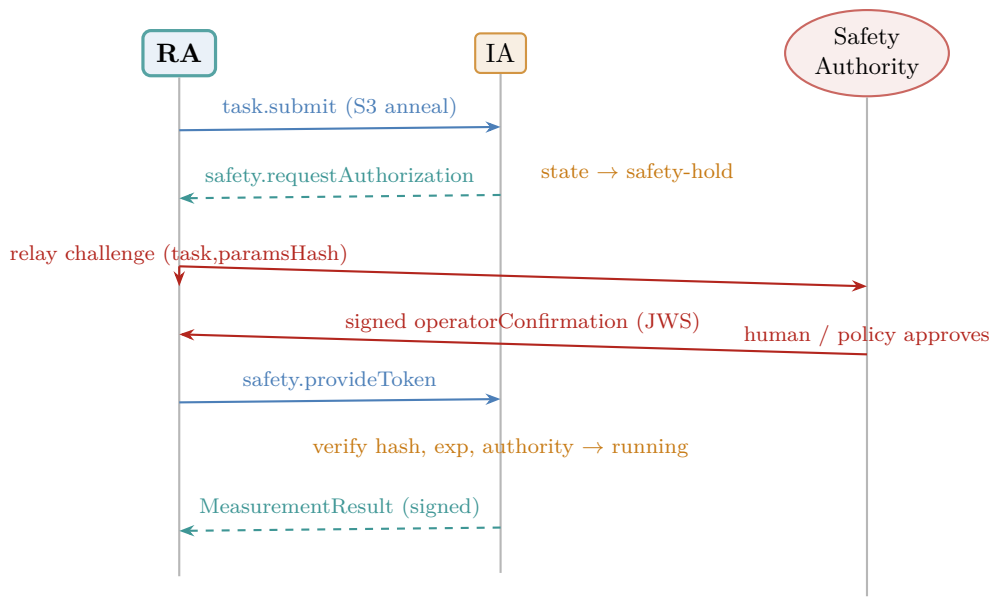
\begin{figure}[t]
\centering
\adjustbox{max width=\linewidth}{%
\begin{tikzpicture}[node distance=22mm]
  % actors
  \node[role,fill=laplight] (ra) {RA};
  \node[box,draw=lapamber!80,fill=lapamber!12,right=34mm of ra] (ia) {IA};
  \node[term,draw=lapred!70,fill=lapred!8,right=34mm of ia,font=\footnotesize] (ho) {Safety\\Authority};
  % lifelines
  \foreach \n in {ra,ia,ho} {\draw[lifeline] (\n.south) -- ++(0,-6.6);}
  \def\y{-0.7}
  \draw[msg] ($(ra.south)+(0,\y)$) -- node[above]{task.submit (S3 anneal)} ($(ia.south)+(0,\y)$);
  \def\y{-1.6}
  \draw[rmsg] ($(ia.south)+(0,\y)$) -- node[above]{safety.requestAuthorization} ($(ra.south)+(0,\y)$);
  \node[font=\scriptsize,lapamber,anchor=west] at ($(ia.south)+(4mm,\y+0.30)$) {state $\to$ safety-hold};
  \def\y{-2.5}
  \draw[msg,lapred] ($(ra.south)+(0,\y)$) -- node[above]{relay challenge (task,paramsHash)} ($(ho.south-|ra.south)+(0,\y)$);
  % the challenge actually goes ra->ho across; draw ra to ho
  \draw[msg,lapred] ($(ra.south)+(0,\y)$) -- node[above,pos=0.75]{} ($(ho.south)+(0,\y)$);
  \def\y{-3.4}
  \node[font=\scriptsize,lapred] at ($(ho.south)+(0,\y+0.25)$) {human / policy approves};
  \draw[msg,lapred] ($(ho.south)+(0,\y)$) -- node[above]{signed operatorConfirmation (JWS)} ($(ra.south)+(0,\y)$);
  \def\y{-4.3}
  \draw[msg] ($(ra.south)+(0,\y)$) -- node[above]{safety.provideToken} ($(ia.south)+(0,\y)$);
  \def\y{-5.2}
  \node[font=\scriptsize,lapamber] at ($(ia.south)+(0,\y+0.25)$) {verify hash, exp, authority $\to$ running};
  \def\y{-6.0}
  \draw[rmsg] ($(ia.south)+(0,\y)$) -- node[above]{MeasurementResult (signed)} ($(ra.south)+(0,\y)$);
\end{tikzpicture}%
}
\caption{The safety-fence handshake for a hazardous (S3) operation. The Instrument Agent
refuses to actuate until it receives an operator-confirmation token whose
\texttt{paramsHash} matches the submitted parameters and whose signature chains to a
recognized Safety Authority. The token authorizes exactly one task with exactly those
parameters.}
\label{fig:safetyseq}
\end{figure}

The \texttt{paramsHash} is the SHA-256 of the capability's parameters in canonical form
(unit-normalized values, lexicographically sorted keys), computed over a string that also
incorporates the \texttt{instr} and \texttt{cap} fields so a token cannot be replayed
against a different instrument or capability. Because the token binds the hash of the exact
parameters that were approved, a Research Agent cannot reuse it to escalate to a different
or more aggressive operation: any change to the parameters, instrument, or capability
changes the bound digest and invalidates the token. The \texttt{jti} nonce makes the token
\emph{single-use}: a conforming Instrument Agent records redeemed \texttt{jti} values until
their \texttt{exp} and rejects any replay, so one human approval authorizes exactly one
execution, not an unbounded series within the validity window. \autoref{fig:safetyseq}
traces the handshake for an S3 operation. This mechanism is what makes ``an autonomous
agent operating hazardous hardware'' a \emph{bounded, auditable} statement rather than an
open-ended one: the agent's authority is exactly the set of task/parameter pairs a human or
policy authority has signed, each redeemable once. We give the full threat model in
\autoref{sec:security}.

\subsection{The MeasurementResult (P4)}
\label{sec:result}
The artifact returned by a completed measurement task is a \texttt{MeasurementResult}
(skeleton in \autoref{lst:result}; complete example with inline data series, raw-artifact
references, and full provenance manifest in \autoref{lst:result-full},
\autoref{app:schemas}). LAP makes four things mandatory that ad-hoc result payloads
routinely omit: every numeric value carries a UCUM \texttt{unit} and the result a QUDT
\texttt{quantityKind}; a \texttt{calibrationRef} links the result to the calibration
under which it was taken; an \texttt{uncertainty} model is stated; and a
\texttt{provenance} block records the exact parameters (with their hash), timing,
instrument firmware, environment, and any operator token used. In addition, when the
producing IA knows facts about the sample's condition that bear on the safety of
\emph{downstream} operations (e.g.\ that a synthesized compound is air-sensitive or has a
thermal history), it records them in a \texttt{sampleCondition} block that the next IA in a
workflow MUST check against its own \texttt{physicalLimits}; this is the protocol carrier
for class~(C) compositional-hazard checking (\autoref{sec:custody}). The whole object is
signed. A result is thus FAIR and reproducible by construction: an agent receiving it,
possibly from a remote laboratory, can interpret it without out-of-band knowledge and
can verify who produced it and under what conditions.

\noindent\begin{minipage}{\linewidth}
\begin{lstlisting}[caption={Skeleton of a \texttt{MeasurementResult} from the powder-scan
   task: a physically typed value (UCUM unit, QUDT quantity kind), an uncertainty model, a
   calibration reference, a provenance digest, and the signature. The complete example is
   given in \autoref{lst:result-full}.},label={lst:result}]
{
  "@type": "lap:MeasurementResult",
  "task": "lap://<lab>/tasks/9f3ab1...",
  "quantityKind": "qudt:Intensity",
  "data": {"intensity": {"unit": "counts", "values": [120, 118, 131]}},
  "uncertainty": {"type": "counting-statistics", "model": "poisson"},
  "calibrationRef": "lap://<lab>/cal/xrd01/2026-05-20",
  "provenance": {"paramsHash": "a91f...",
                 "instrumentFirmware": "D8/3.4.1", "lapVersion": "0.1"},
  "signatures": [{"alg": "ES256", "by": "did:web:<lab>", "jws": "..."}]
}
\end{lstlisting}
\end{minipage}

\subsection{Error model}
\label{sec:errors}
LAP reserves a JSON-RPC error-code range for physical-world failures
(\autoref{tab:errors}). The codes occupy the \texttt{-33xxx} block, placed immediately
below the \texttt{-32768} to \texttt{-32000} band that JSON-RPC~2.0 reserves for its own
predefined and transport-level errors: this keeps LAP's application-level codes cleanly
outside the reserved range while staying numerically adjacent to it. Errors
are designed to be \emph{actionable by an agent}: a
\texttt{ParamOutOfPhysicalLimit} error echoes the offending parameter and its limit so
the agent can re-plan; a \texttt{SafetyAuthorizationRequired} error carries the
authorization challenge directly; a \texttt{ReservationConflict} reports the current
lease holder and expiry.

\begin{table}[t]
\centering\small
\begin{tabular}{@{}rll@{}}
\toprule
\textbf{Code} & \textbf{Name} & \textbf{Meaning}\\
\midrule
$-33001$ & \texttt{ReservationRequired}     & A lease is required before this action.\\
$-33002$ & \texttt{ReservationConflict}      & Instrument held by another lease (holder, expiry returned).\\
$-33003$ & \texttt{LeaseExpired}             & The supplied lease has expired.\\
$-33010$ & \texttt{ParamOutOfPhysicalLimit}  & Parameter violates \texttt{physicalLimits} (offender echoed).\\
$-33020$ & \texttt{SafetyAuthorizationRequired} & S2/S3 op needs a token (challenge attached).\\
$-33021$ & \texttt{SafetyTokenInvalid}       & Token hash/expiry/authority check failed.\\
$-33022$ & \texttt{InterlockTripped}         & A hardware interlock is open.\\
$-33030$ & \texttt{InstrumentFault}          & Device fault; see diagnostics.\\
$-33031$ & \texttt{CalibrationExpired}       & Calibration past \texttt{validUntil}.\\
$-33040$ & \texttt{SampleNotPresent}         & Required sample not loaded.\\
$-33041$ & \texttt{ChainOfCustodyBroken}     & Sample provenance cannot be verified.\\
$-33050$ & \texttt{CapabilityUnsupported}    & No such capability on this instrument.\\
$-33051$ & \texttt{IntentUnresolved}         & \texttt{intent.resolve} could not map the goal (\texttt{reason} in data).\\
$-33060$ & \texttt{BudgetExceeded}            & Lease \texttt{budgetCap} would be exceeded by this task.\\
\bottomrule
\end{tabular}
\caption{The LAP error code range. Codes are designed to let an autonomous agent re-plan
without human intervention by returning the structured context of each failure.}
\label{tab:errors}
\end{table}

\subsection{Security considerations and threat model}
\label{sec:security}
Because LAP actuates physical hardware, its security properties are safety properties. We
state the threat model explicitly. \textbf{Identity and channel:} every LAP entity is
authenticated by mTLS or an OAuth\,2.1 client credential bound to its DID, so message
origin and integrity are assured and an unauthenticated party cannot submit tasks. All
InstrumentCards and MeasurementResults are JWS-signed by the laboratory authority, so a
Research Agent can detect a forged capability advertisement or a tampered result.

\textbf{What the safety fence prevents.} The operator-confirmation token of
\autoref{sec:safety} defends against three concrete attacks. \emph{Parameter escalation:}
because the token binds \texttt{sha256} of the canonical, unit-normalized parameters
together with the instrument and capability identifiers, a Research Agent that obtained
approval for a gentle operation cannot redeem the same token for a more aggressive one; any
change to the request changes the bound digest. \emph{Cross-instrument or cross-capability
replay:} the same binding prevents reusing a token approved for one instrument or method on
another. \emph{Token replay within the validity window:} the \texttt{jti} nonce and the
Instrument Agent's redeemed-token registry make each approval single-use, so one human
decision authorizes exactly one execution.

\textbf{What it does not prevent, and the residual mitigations.} The fence is an
authorization mechanism, not a substitute for a trustworthy authority. A \emph{negligent or
rubber-stamping Safety Authority} that signs without reading defeats it; LAP mitigates but
cannot eliminate this by requiring that the authorization challenge relayed to the human
contain the human-readable parameter values and the capability's declared
\texttt{sideEffects} and hazard class (not the opaque hash alone) so that approval is
informed, and by recommending rate limits and dual-authorization policies for S3 operations.
A \emph{compromised Instrument Agent} is outside the protocol's authority boundary: an IA
that lies about its state or ignores interlocks cannot be constrained by a wire protocol,
which is why physical interlocks remain the last line of defense and the IA's
\texttt{getState} interlock attestation is advisory, not load-bearing. Distinct from a
compromised IA is a \emph{mislabeled InstrumentCard}: a legitimate laboratory, through
negligence or to reduce friction, may sign a card that understates a capability's
\texttt{safetyClass} (declaring a hazardous operation S1) or overstates its
\texttt{physicalLimits}. The card's signature is valid, so signature verification alone does
not catch this, and it is a lower bar for an adversary than compromising a running IA. LAP's
mitigation is to make \texttt{safetyClass} not purely self-asserted: the community
\texttt{instrumentClass} ontology (\autoref{sec:limitations}) SHOULD define, per instrument
class and capability type, a minimum safety class that a conforming card MUST NOT declare
below, and a Federation Registry or Lab Coordinator SHOULD cross-check advertised classes
against this floor and may carry third-party conformance attestations
(\autoref{sec:federation}) that agents weight when selecting a provider. This raises card
misrepresentation from undetectable to auditable, though, like a negligent authority, it
cannot be eliminated by the protocol alone. A \emph{compromised
Research Agent} can still cause harm within its granted authority, but only within it: it
cannot exceed the safety classes and parameter envelopes a human has signed, and every
action it takes is recorded in signed task provenance, which limits and attributes any harm.
\textbf{Operator fatigue} is a real failure mode for autonomous campaigns that generate many
S2/S3 requests; LAP's stance is that the safety class taxonomy should be set so that S3 is
rare by construction, and that standing, scoped policy grants (rather than per-task prompts)
are the right tool for frequent S2 operations, reserving human-in-the-loop confirmation for
hazardous, infrequent actions. \textbf{Denial of service} via reservation
hoarding is bounded by lease expiry and by Lab-Coordinator fair-share and pre-emption
policy (\autoref{sec:reservation}); \textbf{registry trust} and \textbf{Sybil instruments}
are discussed with the federation in \autoref{sec:federation}.

% ============================================================
\section{Discovery, Federation, and a Closed-Loop Walkthrough}
\label{sec:federation}

\subsection{Local and federated discovery}
\label{sec:discovery}
\begin{figure}[t]
\centering
\adjustbox{max width=\linewidth}{%
\begin{tikzpicture}[node distance=14mm and 40mm]
  \node[role,fill=laplight] (ra) {Research\\Agent};
  \node[draw=lapblue!50,thick,fill=laplight,rounded corners,right=of ra,align=center,font=\small]
       (fr) {Federation\\Registry};
  \node[role,right=of fr] (lc) {Remote Lab\\Coordinator};
  \node[box,draw=lapamber!80,fill=lapamber!12,below=16mm of lc] (ia) {Remote\\Instrument Agent};
  \node[box,fill=white,right=10mm of ia,minimum width=14mm] (dev) {\footnotesize PL spec.};

  \draw[flow,lapblue] (ra.north) to[bend left=30] node[midway,above,font=\scriptsize]{1. registry.query (capability)} (fr.north);
  \draw[flow,lapblue] (fr.north) to[bend left=30] node[midway,below=1pt,font=\scriptsize]{2. signed pointers} (ra.north);
  \draw[flow,lapblue,dotted] (lc) -- (fr) node[midway,above,font=\scriptsize]{0. advertise digest};
  \draw[flow,lapteal] (ra.south) to[bend right=20] node[midway,below,font=\scriptsize,align=center]{3. negotiate access\\$\to$ scoped credential} (lc.south);
  \draw[flow,lapamber,very thick] (ra) -- node[pos=0.62,above,sloped,font=\scriptsize\bfseries]{4. direct LAP (reserve, task, stream)} (ia);
  \draw[flow,lapamber,dashed] (ia) -- (dev) node[midway,above,font=\scriptsize]{L0};
  \begin{scope}[on background layer]
    \node[draw=lapgray!40,rounded corners,fill=lapgray!4,fit=(lc)(ia)(dev),inner sep=8pt,
      label={[font=\scriptsize\itshape,lapgray]below:remote trust domain}] {};
  \end{scope}
\end{tikzpicture}%
}
\caption{Federated discovery and access. The registry indexes capability digests and
returns signed pointers (steps 1--2); the Research Agent negotiates a scoped credential
with the owning Lab Coordinator (step 3) and then speaks LAP \emph{directly} to the
remote Instrument Agent (step 4). The registry never proxies control.}
\label{fig:federation}
\end{figure}
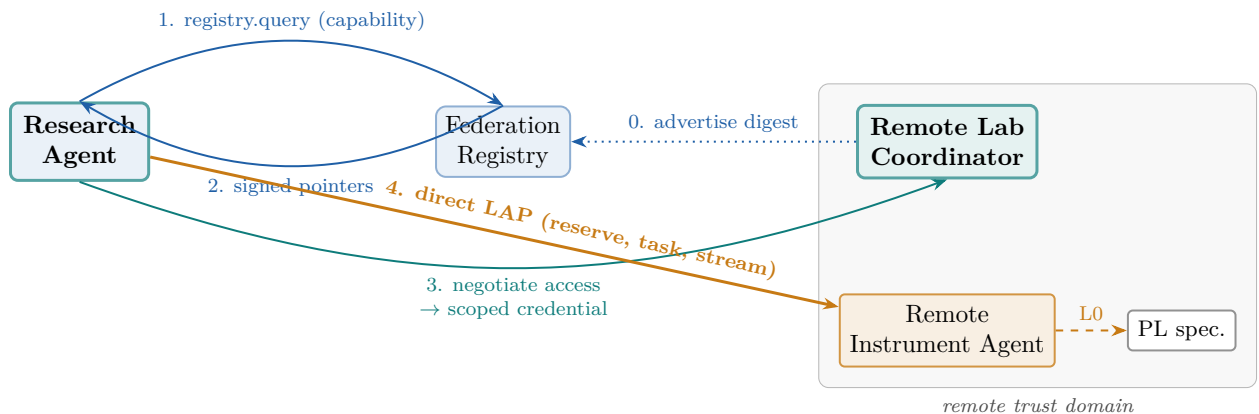

LAP discovery is layered. Within a laboratory, a single Instrument Agent is found at its
\nolinkurl{/.well-known/instrument-card.json}, and a Lab Coordinator lists all of its
instruments at \nolinkurl{/.well-known/lab-directory.json}. Across laboratories, a Lab
Coordinator publishes to one or more Federation Registries with
\texttt{registry.advertise}, sending a \emph{capability digest} (instrument class, key
capability and intent tags, current availability, access policy, and geographic
location) but \emph{not} control endpoints. A Research Agent issues
\texttt{registry.query} (``which instruments can perform \texttt{xrd:PowderDiffraction}
on a sample under 50\,mg, are available this week, and are open to external users?'') and
receives signed pointers to matching laboratories. It then negotiates access with the
owning Lab Coordinator, receives a scoped, time-boxed credential, and talks
LAP \emph{directly} to the remote Instrument Agent (\autoref{fig:federation}). The
registry is a directory, never a control proxy: this keeps latency low and leaves each
laboratory the sole authority over its own instruments. Trust is anchored in
decentralized identifiers; each laboratory is a trust domain, and cross-laboratory access
is an explicit, revocable, scoped credential grant rather than ambient access.

\paragraph{Trust, revocation, and failure modes.} Three properties follow from the
directory-only design. First, the registry is \emph{not a single point of failure for
control}: if it is unavailable or returns a malicious pointer, an agent simply fails to
\emph{discover} a remote instrument, but no instrument can be actuated through the registry,
and every InstrumentCard the agent does retrieve is verified against the issuing
laboratory's signature before it is trusted; so a compromised registry cannot forge
capabilities or redirect control to an impostor. Second, we distinguish two threats that the signature requirement treats differently.
\emph{Identity fraud}---a Sybil attacker advertising instruments under fake
identities---is bounded by the signature requirement and by laboratory-level admission: a
registry entry is only as trustworthy as the laboratory DID that signed it, and registries
may require laboratory identity proofing and carry reputation or accreditation metadata that
agents weight when selecting a provider. \emph{Misrepresentation by a legitimate
laboratory}---a real, correctly-signed lab that nonetheless understates a capability's
\texttt{safetyClass} or overstates its \texttt{physicalLimits}, whether through negligence
or to reduce operator friction---is \emph{not} caught by signature verification, since the
card is genuinely signed. This is the more insidious case, and the federated analog of the
card-misrepresentation threat of \autoref{sec:security}: the mitigation is to make safety
classification not purely self-asserted, by having the registry cross-check advertised
\texttt{safetyClass} values against a community per-\texttt{instrumentClass} floor and carry
optional third-party \emph{conformance attestations} (that a given IA version was verified
to enforce its declared limits) as weighted trust signals that an agent can require for
safety-critical remote operations.
Third, cross-laboratory credentials are \emph{short-lived and scoped} (to specific
capabilities, sample-size envelopes, and time windows) so that revocation is mostly
achieved by expiry; for active revocation the issuing Lab Coordinator maintains a
credential-status endpoint that an Instrument Agent consults for long-running grants. None
of this eliminates the governance problem of \emph{who operates a trusted global registry}
and on what terms, which we leave as an open governance problem (\autoref{sec:discussion}).

\subsection{Sample chain-of-custody}
\label{sec:custody}
Physical samples move between instruments and, in the federated case, between
laboratories. LAP tracks, but does not itself enforce, this movement as a
chain-of-custody on the sample URI. When a Lab Coordinator routes a sample from a
synthesis robot to a diffractometer, it records a custody handoff; the receiving
Instrument Agent's task remains in \texttt{sample-wait} until it attests the sample is
present, and the resulting \texttt{MeasurementResult} carries the sample URI so that the
full physical provenance of a datum is recoverable. The handoff is also where class~(C)
compositional-hazard checking happens: a \texttt{MeasurementResult} carries forward a
\texttt{sampleCondition} block (e.g.\ \texttt{airSensitive}, thermal history) recording what
the producing IA knows about the sample's state, and the Lab Coordinator and the receiving
Instrument Agent MUST check these flags against the receiving instrument's
\texttt{physicalLimits} before the sample is loaded---so that a sequence of
individually-safe steps (a synthesis, then an open-air measurement) cannot compose into a
hazard that no single Instrument Agent could have seen. Breaks in the chain surface as the
\texttt{ChainOfCustodyBroken} error, prompting the Research Agent to reconcile rather than
trust a misattributed sample.

\subsection{A closed-loop campaign, end to end}
\label{sec:walkthrough}
\begin{figure}[t]
\centering
\adjustbox{max width=\linewidth}{%
\begin{tikzpicture}[node distance=24mm]
  \node[role,fill=laplight] (ra) {RA};
  \node[role,right=26mm of ra] (lc) {LC};
  \node[box,draw=lapamber!80,fill=lapamber!12,right=24mm of lc] (syn) {Synth IA};
  \node[box,draw=lapamber!80,fill=lapamber!12,right=24mm of syn] (xrd) {XRD IA};
  \foreach \n in {ra,lc,syn,xrd} {\draw[lifeline] (\n.south) -- ++(0,-7.4);}
  \def\y{-0.6}\draw[msg] ($(ra.south)+(0,\y)$) -- node[above,font=\tiny]{reservation.request} ($(syn.south-|lc.south)+(0,\y)$);
  \draw[msg] ($(ra.south)+(0,\y)$) -- node[pos=0.92,below,font=\tiny]{(robot+XRD)} ($(xrd.south)+(0,\y)$);
  \def\y{-1.4}\draw[msg] ($(ra.south)+(0,\y)$) -- node[above,font=\tiny]{intent.resolve + task.submit (synthesize)} ($(syn.south)+(0,\y)$);
  \def\y{-2.3}\draw[rmsg] ($(syn.south)+(0,\y)$) -- node[above,font=\tiny]{completed: sample URI} ($(ra.south)+(0,\y)$);
  \def\y{-3.1}\draw[msg,lapgray] ($(syn.south)+(0,\y)$) -- node[above,font=\tiny]{custody handoff} ($(xrd.south)+(0,\y)$);
  \node[font=\tiny,lapamber] at ($(xrd.south)+(0,-3.5)$){sample-wait $\to$ present};
  \def\y{-4.1}\draw[msg] ($(ra.south)+(0,\y)$) -- node[above,font=\tiny]{task.submit powder-scan (S1) + task.stream} ($(xrd.south)+(0,\y)$);
  \def\y{-5.0}\draw[rmsg] ($(xrd.south)+(0,\y)$) -- node[above,font=\tiny]{live frames \ldots\ then MeasurementResult} ($(ra.south)+(0,\y)$);
  \def\y{-5.9}\node[font=\tiny,lapteal,align=center] at ($(ra.south)+(2.0,\y)$){RA analyzes $\to$ updates model $\to$ proposes next composition};
  \def\y{-6.6}\draw[msg,lapblue,dashed] ($(ra.south)+(0,\y)$) -- node[above,font=\tiny]{\ldots\ next iteration / federated PL measurement \ldots} ($(xrd.south)+(0,\y)$);
\end{tikzpicture}%
}
\caption{The closed-loop campaign of \autoref{sec:walkthrough} as a message sequence
(local portion). Reservation, intent resolution and confirmation, custody handoff,
streamed measurement, and a typed signed result compose into one iteration of an
autonomous optimization loop; the loop then repeats, optionally reaching across the
federation.}
\label{fig:e2e}
\end{figure}
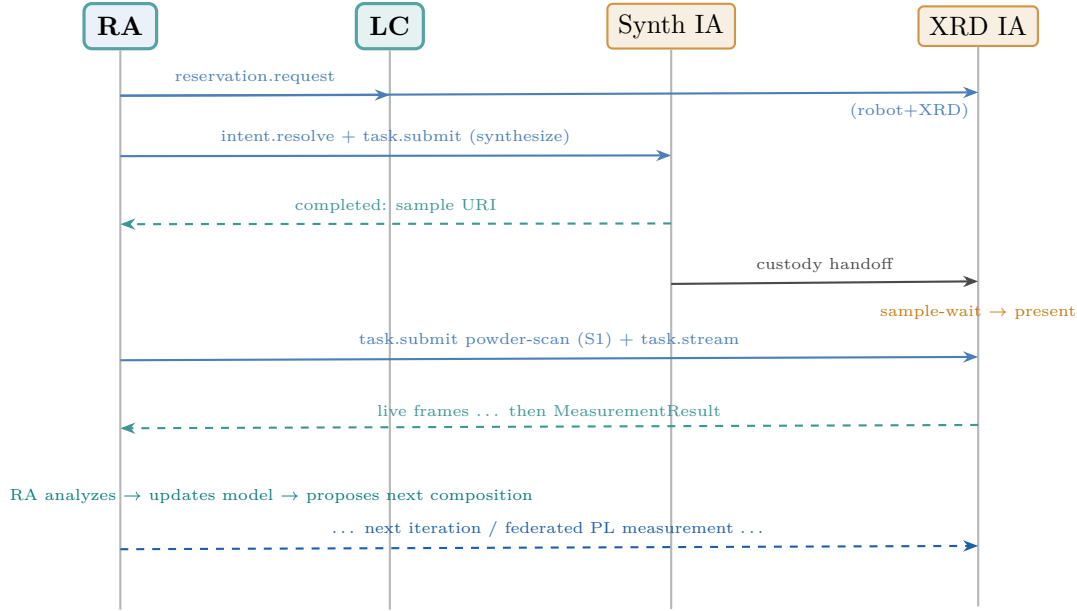

We now run a complete autonomous campaign through the protocol to show how the primitives
compose (\autoref{fig:e2e}). The scenario: a materials Research Agent optimizing a
perovskite composition for stability, mixing local and remote instruments and one
hazardous step.

\begin{enumerate}[leftmargin=1.6em,itemsep=2pt]
\item The RA issues \texttt{registry.query} and finds a local synthesis robot and XRD
  plus a remote photoluminescence (PL) spectrometer.
\item It calls \texttt{reservation.request} to acquire exclusive two-hour leases on the
  synthesis robot and the XRD.
\item It calls \texttt{intent.resolve} (``synthesize Cs$_{0.17}$FA$_{0.83}$PbI$_3$,
  anneal at 100\,$^\circ$C for 10\,min'') and confirms the returned proposal with
  \texttt{task.submit}. The synthesis task runs to \texttt{completed}; its artifact is a
  sample URI.
\item The Lab Coordinator routes the sample to the XRD (custody handoff); the XRD task
  leaves \texttt{sample-wait} once the sample is attested present.
\item The RA submits a powder-scan (class S1) and opens \texttt{task.stream} to watch the
  live pattern; the task completes with a signed, typed \texttt{MeasurementResult}.
\item The RA analyzes phase purity, decides it also needs a PL measurement, negotiates
  federated access to the remote spectrometer, ships the sample, and repeats the
  measurement loop remotely.
\item One refinement step calls for a laser anneal, an S3 operation. The Instrument
  Agent moves the task to \texttt{safety-hold} and issues
  \texttt{safety.requestAuthorization}; a human approves once, the token is bound to that
  exact task and parameters, the anneal executes, and the loop continues.
\item The RA updates its surrogate model and proposes the next composition. Every step is
  auditable through its task provenance.
\end{enumerate}

% ============================================================
\section{Discussion, Limitations, and Comparison}
\label{sec:discussion}

\subsection{Design Rationale}
\label{sec:rationale}

\textbf{Transport: JSON-RPC 2.0 over HTTPS\,+\,SSE.}
LAP mirrors the transport choices of A2A \citep{a2a2025} and MCP \citep{mcp2024}
rather than adopting gRPC as SiLA\,2 does \citep{sila2022standard}.
The reasons are ecosystem alignment, not performance.
JSON-RPC 2.0 over HTTPS is already the wire format of every major agent framework;
an Instrument Agent (IA) that speaks LAP is therefore reachable by any A2A-compatible research agent with zero additional tooling.
gRPC imposes protobuf schema compilation, binary-only debugging, and HTTP/2 firewall issues that are acceptable costs for
SiLA\,2's deterministic-orchestrator use-case but unnecessarily high for the long tail of scientific instruments
whose developers write Python, not generated stubs.
For the small fraction of instruments that require high-frequency sensor streaming (detector frames at kHz rates),
LAP's streaming capability clause permits a WebSocket or gRPC binding alongside the normative SSE path,
keeping the base case simple and the high-performance case possible.

\textbf{Natural language at the edge: confining stochasticity, not guaranteeing safety.}
The central architectural choice in LAP is that natural language \emph{never directly drives hardware}.
The \texttt{intent.resolve} method accepts a free-text goal and returns a \emph{structured proposal}: a fully-typed
\texttt{task.submit} payload with all parameters filled in, which the Research Agent must inspect and
then submit explicitly.
The role of this step is to \emph{confine the stochasticity} of a language model to a proposal layer.
LLMs are stochastic: the same prompt may parse identically worded requests into subtly different parameter values
on different invocations.
By confining this stochasticity to the proposal layer and requiring a deterministic, schema-validated
\texttt{task.submit} to initiate any physical action, LAP ensures that the exact parameters driving hardware
are always inspectable, logged, and replayable.
We are deliberate that this is a \emph{reliability} mechanism, not the protocol's safety guarantee:
\texttt{intent.resolve} is best-effort and optional, and the deterministic \texttt{task.submit} gate is a
\emph{precondition} for, not a substitute for, the safety mechanisms.
LAP separates three kinds of safety and places each where it can be enforced (\autoref{sec:safety}):
\textbf{(A) device-physics safety} (physical-envelope and interlock enforcement) is delegated entirely to the
Instrument Agent as a black box under a normative ``MUST reject out-of-envelope'' obligation;
\textbf{(B) authorization and accountability safety} (human consent for hazardous or irreversible actions) is
standardized at the protocol level as the safety fence with task-bound operator tokens, because it crosses the
human--agent trust boundary and must be uniform across vendors;
and \textbf{(C) workflow- and integrity-level safety} (compositional hazards across steps, expired-calibration
results, and resource/cost runaway) is the concern of the Lab Coordinator together with a few protocol-level
declarations (\texttt{sampleCondition}, mandatory calibration-expiry rejection, and \texttt{budgetCap}).
Reliability follows the same containment logic: a confirmed structured payload can be replayed for reproducibility;
a natural-language string cannot.

\textbf{Reservation and locking as first-class primitives.}
Physical instruments are exclusively-owned, stateful resources: a powder diffractometer cannot run two scans
simultaneously; a liquid-handler arm cannot be in two positions at once.
Neither A2A, MCP, nor SCP \citep{scp2025} provides any reservation or locking primitive.
LAP elevates \texttt{reservation.request}/\texttt{.renew}/\texttt{.release}
to the core method set so that time-bounded exclusive leases are protocol-enforced rather than
left to ad-hoc per-IA convention.
Without this, a multi-agent experiment campaign has no safe way to co-schedule instrument time;
two Research Agents could simultaneously attempt to drive the same instrument into contradictory states.

\textbf{Physically-typed measurement results.}
In existing protocols, tool results are opaque text or arbitrary JSON.
A value of \texttt{25} returned by an MCP server carries no information about whether it is degrees Celsius,
kilopascals, or counts-per-second.
LAP mandates that every numeric value in a \texttt{MeasurementResult} carries a UCUM unit code \citep{ucum},
a QUDT \texttt{quantityKind} URI \citep{qudt}, a \texttt{calibrationRef}, an \texttt{uncertainty} model,
and a signed provenance manifest (parameter hash, timing, firmware version).
This is the minimum information a downstream reasoning agent needs to interpret a measurement correctly,
compose results across instruments, and satisfy FAIR data principles \citep{wilkinson2016fair}.

% -----------------------------------------------------------------------
\subsection{Relationship to Existing Standards}
\label{sec:relationship}

LAP does not compete with SiLA\,2 \citep{sila2022standard}, OPC-UA/LADS \citep{opcua},
or SCPI \citep{scpi}.
These standards operate at \emph{Layer~0} of the LAP stack: the device-abstraction layer that lives
\emph{inside} the Instrument Agent and is entirely the IA's private implementation concern.
An IA wrapping a SiLA\,2 gRPC server translates incoming LAP JSON-RPC calls into outgoing SiLA\,2
Feature invocations; it presents a clean LAP interface upward regardless of what protocol the
physical instrument speaks downward.
Similarly, an IA can wrap an OPC-UA/LADS node, a vendor REST API, or a raw SCPI connection over PyVISA \citep{pyvisa}; the Research Agent sees only LAP.
The relationship is one of \emph{encapsulation}, not replacement.

LAP \emph{composes} with A2A: an Instrument Agent is a valid A2A agent (it publishes an \texttt{InstrumentCard}
at \texttt{/.well-known}, speaks JSON-RPC 2.0 over HTTPS, and carries an A2A-compatible security scheme),
so any A2A-speaking Research Agent can discover and address it, and drive its structured \texttt{task.*} and
\texttt{reservation.*} methods, without LAP-specific client code; the natural-language \texttt{intent.resolve}
path and the physical-world extension fields (reservation, safety class, physical limits) do require a
LAP-aware client to interpret, so ``zero additional tooling'' applies to the structured A2A-shaped surface,
not to the full physical-world semantics.
The \texttt{InstrumentCard} extends A2A's \texttt{AgentCard} pattern with the four physical-world primitives
(capability ontology, reservation profile, safety class, calibration block) that an AgentCard has no fields for.

LAP can coexist with MCP \citep{mcp2024}: an orchestration environment that uses MCP to give
a Research Agent access to computational tools (literature search, DFT calculation, data analysis)
can simultaneously use LAP to give the same agent access to physical instruments.
The two protocols serve orthogonal edges of the agentic graph.

The closest prior work is the Science Context Protocol (SCP) \citep{scp2025},
which extends MCP with a centralized hub, persistent experiment objects, and vendor-agnostic device drivers.
LAP differs in three structural ways: (a) LAP is decentralized: the Lab Coordinator is optional and
federation is peer-to-peer rather than hub-mediated; (b) LAP adds a formal safety fence (S0--S3 classification
plus bound operator tokens) absent from SCP; and (c) LAP mandates physically-typed, calibration-anchored,
uncertainty-bearing results rather than treating result content as opaque.
Table~\ref{tab:comparison} summarizes the full comparison.

\begin{table*}[t]
\centering
\small
\setlength{\tabcolsep}{4pt}
\adjustbox{max width=\linewidth}{%
\begin{tabular}{p{2.7cm}p{2.1cm}p{1.9cm}p{2.0cm}p{1.9cm}p{2.5cm}}
\hline
\textbf{Dimension} & \textbf{SiLA\,2} & \textbf{MCP} & \textbf{A2A} & \textbf{SCP} & \textbf{LAP} \\
\hline
Primary edge & sw-orch.\ $\leftrightarrow$ device & agent $\leftrightarrow$ tool & agent $\leftrightarrow$ agent & agent $\leftrightarrow$ tool (hub) & \textbf{agent $\leftrightarrow$ instr.} \\[3pt]
Discovery & mDNS, static & host config & AgentCard \texttt{/.well-known} & hub registry & InstrumentCard + federation \\[3pt]
NL / intent & none & tool descriptions & skills (text) & tool descs & \texttt{intent.resolve} \\[3pt]
Physical-qty typing & basic types & none & none & partial & QUDT/UCUM \\[3pt]
Reservation & none & none & none & session & first-class leases \\[3pt]
Safety fence & none & none & none & none & S0--S3 + tokens \\[3pt]
Calibration / uncert. & data only (AnIML) & none & none & partial & mandatory \\[3pt]
Cross-lab federation & none & none & partial & hub & registry + scoped VC \\[3pt]
Transport & gRPC & JSON-RPC & JSON-RPC / gRPC & JSON-RPC & JSON-RPC + SSE \\[3pt]
Running impl. & mature, multi-vendor & wide & growing & deployed (hub) & \textbf{none (design)} \\
\hline
\end{tabular}%
}
\caption{Comparison of LAP with related protocols. SiLA\,2, OPC-UA/LADS, EPICS, and SCPI live at Layer~0
  (inside the IA); MCP, A2A, and SCP operate at higher layers and are \emph{complementary} to LAP, not competing.
  The final row states LAP's status plainly: it is a specification, and unlike the comparators it has no
  implementation yet; the reference prototype of \autoref{sec:adoption} is its immediate next step.}
\label{tab:comparison}
\end{table*}

% -----------------------------------------------------------------------
\subsection{Limitations}
\label{sec:limitations}

\begin{enumerate}

\item \textbf{Specification-first; empirical validation is the next phase.}
LAP v0.1 is, by design, a protocol and schema specification rather than a finished,
benchmarked system, following the same specification-first path that MCP and A2A took before their
ecosystems and implementations formed around a public, neutrally governed standard. Its
benefits are at this stage established by architectural argument rather than by production
deployment, multi-site conformance data, or controlled user studies. Closing that
gap, starting with the reference implementation and conformance suite described in
\autoref{sec:adoption} and progressing toward multi-laboratory deployment, is the explicit
goal of the open, community-built effort this paper is intended to seed.

\item \textbf{Real-time hard-deadline control is out of scope.}
LAP targets \emph{task-level orchestration}: submitting a 20-minute XRD scan, monitoring its progress,
and retrieving the signed result.
It is explicitly not designed for microsecond-level closed-loop servo control, PID loops,
or any application where the control latency of JSON-RPC over HTTPS is unacceptable.
Instruments with hard real-time requirements (e.g., ultrafast pump-probe spectroscopy, fast feedback for
scanning probe microscopy) must implement their inner control loop below Layer~0,
outside the LAP scope entirely.
LAP may invoke such instruments at the task level but cannot replace their real-time firmware.

\item \textbf{The safety fence assumes a trustworthy operator authority.}
The S2/S3 operator-confirmation token mechanism (Section~7 of the design specification) binds a
specific human approval to a specific task and parameter hash.
This prevents a Research Agent from reusing a token for a different operation.
However, it does not prevent a \emph{malicious or negligent Safety Authority} from signing tokens for
hazardous operations they should not approve, nor does it prevent a compromised IA from ignoring tokens
entirely.
LAP provides protocol-level hooks for auditable human oversight; it cannot substitute for
institutional safety governance, physical interlocks, and operator training.
Two related residual risks follow from delegating device-physics safety to the IA: a legitimate laboratory may
sign an InstrumentCard that \emph{understates} a capability's safety class, which LAP raises from undetectable to
auditable (but does not eliminate) via a community-defined per-\texttt{instrumentClass} safety-class floor and
registry conformance attestations (\autoref{sec:security}); and hazards visible only \emph{across} steps---an
individually-safe sequence that is jointly unsafe, or a campaign that consumes unbounded cost---are addressed not
by the fence but by Lab-Coordinator workflow checks and the \texttt{sampleCondition} and \texttt{budgetCap}
declarations, which presuppose a Lab Coordinator and therefore do not protect the minimal single-IA topology.

\item \textbf{The capability ontology (\texttt{instrumentClass}) is under-specified.}
The \texttt{InstrumentCard} includes an \texttt{instrumentClass} field (e.g.,
\texttt{xrd:PowderDiffractometer}) intended to enable semantic capability matching across labs.
In LAP v0.1, the vocabulary for this field is not formally defined; we reference it by example.
Without a community-curated, versioned ontology of instrument classes, two labs using different
\texttt{instrumentClass} strings for nominally equivalent instruments will be undiscoverable to
each other.
Developing and governing this ontology requires sustained community effort that has not yet begun.

\item \textbf{Chain-of-custody for physical samples depends on real-world logistics LAP cannot enforce.}
LAP tracks sample provenance via signed URI handles and records handoffs between Instrument Agents in
an audit log.
It does \emph{not} enforce that the physical sample arrives at the correct instrument at the correct time,
that it has not been mislabeled in transit, or that the sample a technician loads onto the instrument
is the one the URI claims.
Bridging the digital chain-of-custody record to the physical world requires barcodes, RFID, robotics,
and human procedures that are entirely outside the protocol's scope.

\item \textbf{\texttt{intent.resolve} reliability is best-effort and LLM-dependent.}
The natural-language grounding path (where an IA's \texttt{intent.resolve} method interprets a
free-text goal and proposes a typed \texttt{task.submit} payload) depends on the quality of the
underlying language model and the completeness of the \texttt{intentTags} in the \texttt{InstrumentCard}.
There is no guarantee that the proposal correctly reflects the scientist's intent,
particularly for ambiguous requests, instrument-specific jargon, or edge cases not covered by the
capability's \texttt{intentTags}.
The design requirement that a Research Agent must explicitly confirm the proposal before submission
mitigates but does not eliminate this risk.

\item \textbf{Federation trust and credentialing at global scale is an unsolved governance problem.}
LAP's federation design sketches a model in which labs issue scoped, time-boxed credentials to
external Research Agents via DID-based identity and verifiable credentials.
At small scale (a consortium of trusted partner labs), this model is plausible.
At the scale of a ``scientific internet of instruments'' spanning thousands of institutions across
jurisdictions with differing legal frameworks, data sovereignty requirements, and biosecurity regulations,
the governance infrastructure required to issue, validate, revoke, and audit cross-lab credentials
does not exist and is not a problem LAP can solve unilaterally.
Federated identity at this scale is an open research problem in its own right.

\end{enumerate}

% -----------------------------------------------------------------------
\subsection{Path to Adoption}
\label{sec:adoption}

LAP is designed for incremental adoption with minimal migration burden. An existing instrument
becomes an Instrument Agent through a thin software adapter that maps LAP JSON-RPC onto the
device's native layer (a SiLA\,2 gRPC server, a vendor SDK, or a raw SCPI connection) with no
change to the instrument or its drivers.

Two adoption stages follow from this. In the near term, the burden falls on integrators: a lab,
a consortium, or an open-source community wraps an existing instrument in a thin Instrument Agent,
requiring no cooperation from the original manufacturer. In the longer term, the same L0
encapsulation that keeps a vendor SDK private below the protocol lets the \emph{manufacturer ship
the Instrument Agent natively}, as a product feature, exposing a uniform LAP face without
disclosing or altering proprietary firmware. Because an Instrument Agent is a valid A2A agent, a
LAP-native instrument is immediately addressable by every agent framework in the ecosystem with no
per-customer integration, amortizing the adapter cost once, at the manufacturer, rather than
repeatedly, at every deployment, exactly as device-class standards did for USB peripherals and IPP
did for network printers. This two-stage path is deliberately not vendor-gated: stage one is
actionable today without manufacturer buy-in, and stage two follows once the standard demonstrates
pull.

The immediate next step, and the smallest experiment that de-risks the whole design, is an open
reference implementation and conformance suite: the four primitives can be exercised end to end on
a single laptop, with no hardware, by a \emph{simulated XRD Instrument Agent} (serving an
InstrumentCard, streaming synthetic frames over SSE, returning a signed \texttt{MeasurementResult},
and exposing one S1 and one S3 capability), a minimal \emph{Safety Authority} that issues the
parameter-hash-bound operator-confirmation token, a scripted \emph{Research Agent} that drives the
full discover\,$\to$\,reserve\,$\to$\,resolve\,$\to$\,submit\,$\to$\,safety-hold flow, and a
\texttt{pytest} suite asserting every state transition and that altering any approved parameter
invalidates the token. We report this as planned work, not completed evaluation
(\autoref{sec:limitations}). Beyond it lie an open, community-curated InstrumentCard registry, which
would also seed the \texttt{instrumentClass} ontology, and eventually stewardship under a neutral
multi-stakeholder body, the governance path that MCP and A2A followed.

% ============================================================
\FloatBarrier
\section{Conclusion}
\label{sec:conclusion}
We have argued that autonomous science is limited less by the intelligence of its
agents than by the absence of a protocol for the edge where intelligence meets matter.
The Model Context Protocol gave agents a standard way to reach tools, and the
Agent2Agent protocol gave agents a standard way to reach each other; the Lab Agent
Protocol completes the triangle by giving agents a standard, safe, and physically
faithful way to reach instruments. LAP's contribution is not a new transport or a new
data format (it reuses JSON-RPC, JSON Schema, QUDT/UCUM, W3C WoT Thing
Descriptions, and the A2A discovery and task-lifecycle patterns) but the combination of
four primitives that the physical world demands and that no prior agent protocol
provides: the InstrumentCard, exclusive reservation, the safety-fence
handshake with task-bound operator tokens, and the calibration- and
uncertainty-anchored MeasurementResult. We specified these end to end and showed how
they compose into a closed autonomous research loop and into a federated
Scientific Internet of Instruments. The design is a starting point, not a finished
standard: it has no large-scale implementation or empirical evaluation yet, its
capability ontology needs community curation, and its federation raises governance and
safety questions that a specification can make addressable but cannot resolve alone. We
offer LAP as a concrete basis for that community effort, in the open-governance spirit
the agent-protocol community has adopted for A2A and MCP.

\section*{Acknowledgments}
We thank several senior engineers at Shiyanjia Lab for their collaboration and
support throughout this work.

\bibliographystyle{plainnat}
\bibliography{references}

\clearpage
\appendix
\section{Supplementary Material: Complete Schema Examples}
\label{app:schemas}

This appendix gives the full versions of the two central LAP artifacts whose abbreviated
skeletons appear in the main text: the \textbf{InstrumentCard} (\autoref{sec:card},
\autoref{lst:card}) and the \textbf{MeasurementResult} (\autoref{sec:result},
\autoref{lst:result}). They are reproduced here in full so that the main text can stay
focused on the protocol's structure while a reader who wants the complete field-level
detail has an authoritative reference.

\begin{lstlisting}[caption={Complete (abridged) InstrumentCard for a powder XRD instrument.
   The card declares one capability (\texttt{powder-scan}) with physically typed parameters,
   a safety class, physical limits, and intent tags, plus live telemetry streams and a
   calibration block. Fields shared with an A2A AgentCard (transport, security) are
   retained; the lab-specific fields are LAP additions.},label={lst:card-full}]
{
  "@context": ["https://www.w3.org/2022/wot/td/v1.1",
               "https://<lap-namespace>/context/v0.1"],
  "@type": "lap:InstrumentCard",
  "id": "lap://<lab>/instruments/xrd-bruker-d8-01",
  "lapVersion": "0.1",
  "title": "Bruker D8 Advance Powder XRD",
  "instrumentClass": "xrd:PowderDiffractometer",
  "vendor": {"name": "Bruker", "model": "D8 Advance", "serial": "DA-2231"},
  "location": {"lab": "lap://<lab>/", "room": "B-114", "geo": [42.36, -71.09]},
  "lapProfile": {"streaming": true, "reservation": true,
                 "safetyFence": true, "federation": true,
                 "intentResolve": true},
  "interfaces": [{"protocolBinding": "lap-jsonrpc",
                  "url": "https://<host>/lap",
                  "preferredTransport": "https+sse"}],
  "securitySchemes": {"mtls": {"scheme": "mtls"}},
  "securityRequirements": ["mtls"],
  "capabilities": [{
    "id": "powder-scan",
    "name": "Powder theta-2theta scan",
    "intentTags": ["measure crystal structure", "phase identification",
                   "acquire diffraction pattern"],
    "inputSchema": {
      "type": "object",
      "required": ["twoThetaStart", "twoThetaEnd", "stepSize"],
      "properties": {
        "twoThetaStart": {"$ref": "lap:Quantity", "unit": "deg",
                          "minimum": 5, "maximum": 120},
        "twoThetaEnd":   {"$ref": "lap:Quantity", "unit": "deg",
                          "minimum": 5, "maximum": 120},
        "stepSize":      {"$ref": "lap:Quantity", "unit": "deg", "minimum": 0.001},
        "timePerStep":   {"$ref": "lap:Quantity", "unit": "s", "default": 0.5}}},
    "outputSchema": {"$ref": "lap:MeasurementResult"},
    "safetyClass": "S1",
    "reversible": true,
    "estimatedDuration": {"value": 1200, "unit": "s"},
    "consumesSample": true,
    "sideEffects": ["X-ray shutter opens"],
    "physicalLimits": {"maxPower": "2.2 kW",
                       "interlocks": ["shutter", "doorClosed"]},
    "operationalCost": {"currency": "USD", "perExecution": 8.0}
  }],
  "streams": [
    {"id": "live-pattern", "encoding": "application/lap-frames+ndjson",
     "rateHz": 2, "quantity": {"unit": "counts"}},
    {"id": "detector-temp", "quantity": {"unit": "degC"}, "rateHz": 0.1}],
  "calibration": {"lastCalibrated": "2026-05-20T08:00:00Z",
                  "validUntil": "2026-06-20", "standard": "NIST SRM 1976c",
                  "calibrationRef": "lap://<lab>/cal/xrd01/2026-05-20"},
  "signatures": [{"alg": "ES256", "by": "did:web:<lab>", "jws": "..."}]
}
\end{lstlisting}

\begin{lstlisting}[caption={A complete \texttt{MeasurementResult} from the powder-scan task.
   Values are physically typed (UCUM units, QUDT quantity kind), calibration-anchored,
   uncertainty-bearing, and wrapped in a signed provenance manifest. Bulk data is
   referenced by hashed artifact; small series may be inlined.},label={lst:result-full}]
{
  "@type": "lap:MeasurementResult",
  "task": "lap://<lab>/tasks/9f3ab1...",
  "capability": "powder-scan",
  "instrument": "lap://<lab>/instruments/xrd-bruker-d8-01",
  "sample": "lap://<lab>/samples/2026-0531-A7",
  "quantityKind": "qudt:Intensity",
  "data": {
    "inline": {
      "twoTheta":  {"unit": "deg",    "values": [5.00, 5.02, 5.04]},
      "intensity": {"unit": "counts", "values": [120, 118, 131]}},
    "artifacts": [{"role": "raw", "mediaType": "application/x-bruker-raw",
                   "url": "https://<host>/blob/9f3ab1.raw",
                   "sha256": "5e1c..."}]},
  "uncertainty": {"type": "counting-statistics", "model": "poisson"},
  "sampleCondition": {"airSensitive": false, "thermalHistory": "as-synthesized"},
  "calibrationRef": "lap://<lab>/cal/xrd01/2026-05-20",
  "provenance": {
    "params": {"twoThetaStart": 5, "twoThetaEnd": 80, "stepSize": 0.02},
    "paramsHash": "a91f...", "resolveRef": "lap://<lab>/resolutions/res-7f3a",
    "startedAt": "2026-05-31T13:02:10Z",
    "endedAt": "2026-05-31T13:22:01Z", "operatorToken": null,
    "instrumentFirmware": "D8/3.4.1", "lapVersion": "0.1",
    "environment": {"roomTemp": {"value": 21.4, "unit": "degC"}}},
  "signatures": [{"alg": "ES256", "by": "did:web:<lab>", "jws": "..."}]
}
\end{lstlisting}

\end{document}